\documentclass{article}

\usepackage{times}
\usepackage{graphicx} 
\usepackage{subcaption}

\usepackage{comment}
\usepackage{xspace}
\usepackage{amsmath}
\usepackage{amsthm}
\usepackage{amsfonts}
\usepackage{multirow}

\newtheorem{remark}{Remark}

\usepackage{natbib}
\usepackage{fontawesome}
\usepackage{algorithm}
\usepackage{algorithmic}
\usepackage{color}

\usepackage{hyperref}

\newcommand{\methodname}[0]{Mix \& Match\xspace}
\newcommand{\methodshort}[0]{M\&M\xspace}
\newcommand{\methodsymbol}[0]{\mathrm{mm}}

\newcommand{\dkl}[0]{\mathrm{D}_\mathrm{KL}}

\usepackage[accepted]{arxiv}
\usepackage{titlesec}

\icmltitlerunning{\methodshort -- Agent Curricula for RL}

\begin{document} 

\twocolumn[
\icmltitle{\methodname{}  -- Agent Curricula for Reinforcement Learning}




\icmlsetsymbol{equal}{*}

\begin{icmlauthorlist}
\icmlauthor{Wojciech Marian Czarnecki}{equal,goo}
\icmlauthor{Siddhant M. Jayakumar}{equal,goo}
\icmlauthor{Max Jaderberg}{goo}
\icmlauthor{Leonard Hasenclever}{goo}
\icmlauthor{Yee Whye Teh}{goo}
\icmlauthor{Simon Osindero}{goo}
\icmlauthor{Nicolas Heess}{goo}
\icmlauthor{Razvan Pascanu}{goo}

\end{icmlauthorlist}

\icmlaffiliation{goo}{DeepMind, London, UK}

\icmlcorrespondingauthor{Wojciech M. Czarnecki}{lejlot@google.com}
\icmlcorrespondingauthor{Siddhant M. Jayakumar}{sidmj@google.com}

\icmlkeywords{reinforcement learning, curriculum learning}

\vskip 0.3in
]



\printAffiliationsAndNotice{\icmlEqualContribution} 

\begin{abstract}

We introduce \methodname (\methodshort) -- a training framework designed to facilitate rapid and effective learning in RL agents, especially those that would be too slow or too challenging to train otherwise.
The key innovation is a procedure that allows us to automatically form a \emph{curriculum over agents}. Through such a curriculum we can progressively train more complex agents by, effectively, bootstrapping from solutions found by simpler agents.
In contradistinction to typical curriculum learning approaches, we do not gradually modify the tasks or environments presented, but instead use a process to gradually alter how the policy is represented internally.
We show the broad applicability of our method by demonstrating significant performance gains in three different experimental setups: 
(1) We train an agent able to control more than 700 actions in a challenging 3D first-person task; using our method to progress through an action-space curriculum we achieve both faster training and better final performance than one obtains using traditional methods.
(2) We further show that \methodshort can be used successfully to progress through a curriculum of architectural variants defining an agents internal state. 
(3) Finally, we illustrate how a variant of our method can be used to improve agent performance in a multitask setting.

\end{abstract}

\begin{figure}[t] \centering
\includegraphics[width=0.4\textwidth]{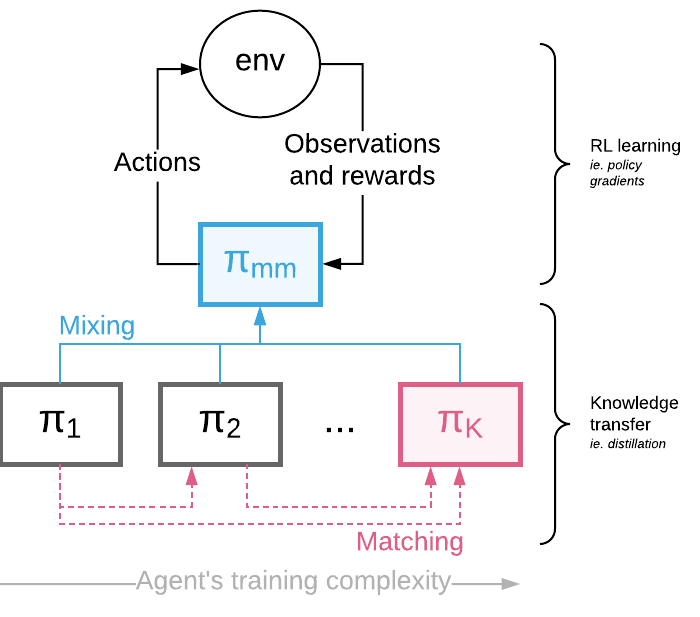}
\caption{Scheme of \methodname -- each box represents a policy. The blue $\pi_\methodsymbol$ is the control policy, optimised with the true RL objective. White boxes represent policies used for the curriculum, while the red policy is the final agent.}%
\label{fig:mm}
\end{figure}

\section{Introduction}

The field of deep reinforcement learning has seen significant advances in the recent past. Innovations in environment design have led to a range of exciting, challenging and visually rich 3D worlds~\citep[e.g.][]{dmlab, kempka2016vizdoom, brockman2016openai}. These have in turn led to the development of more complex agent architectures and necessitated massively parallelisable policy gradient and Q-learning based RL algorithms~\citep[e.g.][]{mnih2016asynchronous, impala}. 
While the efficacy of these methods is undeniable, the problems we consider increasingly require more powerful models, complex action spaces and challenging training regimes for learning to occur.

Curriculum learning is a powerful instrument in the deep learning toolbox \citep[e.g.][]{bengio2009,Graves17}. In a typical setup, one trains a network sequentially on related problems of increasing difficulty, with the end goal of maximizing final performance on a desired task. 
However, such task-oriented curricula pose some practical difficulties for reinforcement learning.
For instance, they require a certain understanding of and control over
the generation process of the environment, such that simpler variants of the task can be constructed.
And in situations where this is possible, it is not always obvious how to construct useful curricula -- simple intuitions from learning in humans do not always apply to neural networks.
Recently 
\citep[e.g.][]{Sutskever14, Graves17} proposes randomised or automated curricula to circumvent some of these issues 
with some success.
In this paper, instead of curricula over task variants we consider an alternate formulation -- namely a curriculum over variants of \emph{agents}. We are interested in training a single final agent, and in order to do so we leverage a series of intermediate agents that differ structurally in the way in which they construct their policies (Fig.~\ref{fig:mm}).
Agents in such curricula are not arranged according to architectural complexity, but rather training complexity.
While these complexity measures are often aligned, they are sometimes orthogonal (\emph{e.g.} it is often faster to train a complex model on two distinct tasks, than a simpler model on them jointly).
In contrast to a curriculum over tasks, our approach can be applied to a wide variety of problems where we do not have the ability to modify the underlying task specification or design. However, in domains where traditional curricula are applicable, these two methods can be easily combined.

The primary contribution of this work is thus to motivate and provide a principled approach for training with curricula over agents.
\subsection*{\methodname: An overview}
In the \methodname{} framework, we treat multiple agents of increasing learning complexity as one \emph{\methodshort agent}, which 
acts with a \emph{mixture} of policies from its constituent agents (Fig.~\ref{fig:mm}).
Consequently it can be seen as an ensemble or a mixture of experts  agent, which is used solely for purpose of training.
Additionally, knowledge transfer (\emph{i.e.} distillation) is used such that we encourage the complex agents to \emph{match} the simpler ones early on.
The mixing coefficient is controlled such that ultimately only the complex, target agent is used for generating experience.
Note that we consider the complexity of an agent not just in terms of the depth or size of its network (see Section 4.2), but with reference to the difficulty in training it from scratch (see Section 4.1 and 4.3).
We also note that while analogous techniques to ours can be applied to train mixtures of experts/policies (\emph{i.e.} maximising performance across agents), this is not the focus of the present research; here our focus is to train a final target agent.

Training with the \methodname{} framework confers several potential benefits -- for instance performance maximisation (either with respect to score or data efficiency), or enabling effective learning in otherwise hard-to-train models. 
And with reference to this last point, \methodshort might be particularly beneficial in settings where real world constraints (inference speed, memory) demand the use of certain specific final models.

\section{Related work}

Curriculum learning is a long standing idea in machine learning, with mentions as early as 
the work of Elman \citep{elman93}. 
In its simplest form, pretraining and finetuning is a form of curriculum, widely explored \citep[\emph{e.g.}][]{Simonyan14c}. 
More explicitly, several works look at the importance of a curriculum for neural networks  \citep[\emph{e.g.}][]{bengio2009}. In many works, this focus is on constructing a sequence of tasks of increasing difficulty. More recent work \cite{Graves17, Sutskever14} however looks at automating task selection or employing a mixture of difficulties at each stage in training. We propose to extend this idea and apply it instead to training agents in curricula -- keeping in spirit recent ideas of mixtures of tasks (here, models).

The recent work on Net2Net \citep{Chen2015Net2NetAL} proposes a technique to increase the capacity of a model without changing the underlying function it represents. 
In order to achieve this, the architectures have to be supersets/subsets of one another and be capable of expressing identity mappings.
Follow-up work \citep{pmlr-v48-wei16} extends these ideas further. Both approaches can be seen as 
implicitly constructing a form of curriculum over 
the architecture, as a narrow architecture is first trained, then morphed into a wider one. 

Related to this idea is the concept of knowledge transfer or distillation \cite{hinton2015distilling,BaDistill14} -- a technique for transferring the functional behaviour 
of a network into a different model, regardless 
of the structure of the target or source networks.
While initially proposed for model compression \citep{buciluǎ2006model, BaDistill14}
, in \cite{ParisottoBS15,rusu-distillation-2015}
distillation is used to compress multiple distinct policies into a single one. 
Distral~\citep{distral} instead focuses on learning independent policies, which use co-distilled centralised agent as a communication channel.

Our work borrows and unifies several of these threads with a focus on online, end-to-end training of model curricula from scratch.

\section{Method details}

We first introduce some notation to more precisely describe our framework.
Let us assume we are given a sequence of trainable agents\footnote{
For simplicity of notation we are omitting time dependence of all random variables, however we do consider a time-extended setting. In particular, when we talk about policy of agent $i$ we refer to this agent policy at given time (which will change in the next step).}
(with corresponding policies $\pi_1, ..., \pi_K$, each parametrised with some $\theta_i \subset \theta$ -- which can share some parameters) ordered according to the complexity of interest 
(\emph{i.e.} $\pi_1$ can be a policy using a tiny neural network while $\pi_K$ the very complex one). 
The aim is to train $\pi_K$, while all remaining agents are there to induce faster/easier learning.
Furthermore, let us introduce the categorical random variable $c \sim \mathrm{Cat}(1,...,K|\alpha)$ (with probability mass function $p(c=i) = \alpha_i$) which will be used to select a policy at a given time:
$$
\pi_{\methodsymbol}(a| s) = \sum_{i=1}^K \alpha_i \pi_i(a|s).
$$
The point of \methodname is to allow curriculum learning, consequently we need the probability mass function (pmf) of $c$ to be changed over time. Initially the pmf should have $\alpha_1=1$ and near the end of training $\alpha_K=1$ thus allowing the curriculum of policies from simple $\pi_1$ to the target one $\pi_K$. Note, that $\alpha_i$ has to be adjusted in order to control learning dynamics and to maximise the whole learning performance, rather than immediate increase. 
Consequently it should be trained in a way which maximises long lasting increase of performance (as opposed to gradient based optimisation which tends to be greedy and focus on immediate rewards).

We further note that mixing of policies is necessary but not sufficient to obtain \emph{curriculum learning} -- even though (for non dirac delta like $c$) gradients always flows through multiple policies, there is nothing causing them to actually share knowledge. In fact, this sort of mixture of experts is inherently competitive rather than cooperative~\cite{jacobs1991adaptive}. In order to address this issue we propose using a distillation-like cost $D$, which will align the policies together.
$$
\mathcal{L}_{\methodsymbol}(\theta) = \sum_{i,j=1}^K \mathrm{D}( \pi_i(\cdot|\cdot, \theta_i), \pi_j(\cdot|\cdot, \theta_j), i, j, \alpha).
$$
The specific implementation of the above cost will vary from application to application. In the following sections we look at a few possible approaches.  

The final optimisation problem we consider is just a weighted sum of the original $\mathcal{L}_\mathrm{RL}$ loss (\emph{i.e.} A3C~\cite{mnih2016asynchronous}), applied to the control policy $\pi_{\methodsymbol}$ and the knowledge transfer loss:
$$
\mathcal{L}(\theta) = \mathcal{L}_\mathrm{RL}(\theta|\pi_{\methodsymbol}) + \lambda \mathcal{L}_{\methodsymbol}(\theta).
$$
We now describe in more detail each module required to implement \methodname, starting with policy mixing, knowledge transfer and finally $\alpha$ adjustment.

\subsection{Policy mixing}

There are two equivalent views of the proposed policy mixing element -- one can either think about having a categorical selector random variable $c$ described before, or an explicit mixing of the policy. The expected gradients of both are the same:
\begin{equation*}
\begin{aligned}
&\mathbb{E}_{i \sim c} \left [ \nabla_\theta \pi_i(a|s,\theta) \right ] = \sum_{i=1}^K \alpha_i \nabla_\theta  \pi_i(a|s,\theta)  =\\
&=\nabla_\theta \sum_{i=1}^K \alpha_i  \pi_i(a|s,\theta) = \nabla_\theta \pi_{\methodsymbol}(a|s,\theta),
\end{aligned}
\end{equation*}
however, if one implements the method by actually sampling from $c$ and then executing a given policy, the resulting single gradient update will be different than the one obtained from explicitly mixing the policy. From this perspective it can be seen as a Monte Carlo estimate of the mixed policy, thus for the sake of variance reduction we use explicit mixing in all experiments in this paper.

\subsection{Knowledge transfer}
For simplicity we consider the case of $K=2$, but all following methods have a natural extension to an arbitrary number of policies. Also, for notational simplicity we drop the dependence of the losses or policies on $\theta$ when it is obvious from context.

Consider the problem of ensuring that final policy $\pi_2$ matches the simpler policy $\pi_1$, while having access to samples from the control policy, ${\pi_\methodsymbol}$.
For simplicity, we define our \methodshort loss over the trajectories directly, similarly to the unsupervised auxiliary losses~\cite{unreal, mirowski2016learning}, thus we put:
\begin{equation}
\label{eq:mm}
\mathcal{L}_\methodsymbol(\theta) = \frac{1-\alpha}{|S|} \sum_{s \in S} \sum_{t=1}^{|s|}  \dkl(\pi_1(\cdot|s_t) \| \pi_2(\cdot|s_t)),
\end{equation}
and trajectories ($s \in S$) are sampled from the control policy. The $1-\alpha$ term is introduced so that the distillation cost disappears when we switch to $\pi_2$.\footnote{It can also be justified as a distillation of mixture policy, see Appendix for derivation.}
This is similar to the original policy distillation~\cite{rusu-distillation-2015}, however here the control policy is mixture of $\pi_2$ (the student) and $\pi_1$ (the teacher). 

One can use a memory buffer to store $S$ and ensure that targets do not drift too much~\cite{ross2011reduction}.
In such a setting, under reasonable assumptions one can prove the convergence of $\pi_2$ to $\pi_1$ given enough capacity and experience.
\begin{remark}
Lets assume we are given a set of $N$ trajectories from some predefined mix $\pi_\methodsymbol = (1-\alpha)\pi_1 + \alpha \pi_2$ for any fixed $\alpha \in (0,1)$ and a big enough neural network with softmax output layer as $\pi_2$.
Then in the limit as $N \rightarrow \infty$, the minimisation of Eq. \ref{eq:mm} converges to $\pi_1$ if the optimiser used is globally convergent when minimising cross entropy over a finite dataset.
\end{remark}
\begin{proof}
Given in the appendix. 
\end{proof}

In practice we have found that minimising this loss in an online manner (\emph{i.e.} using the current on-policy trajectory as the only sample for Eq.~\eqref{eq:mm}) works well in the considered applications.

\begin{figure*}[ht] \centering
\begin{subfigure}[b]{0.3\textwidth}
\centering
\includegraphics[height=5cm]{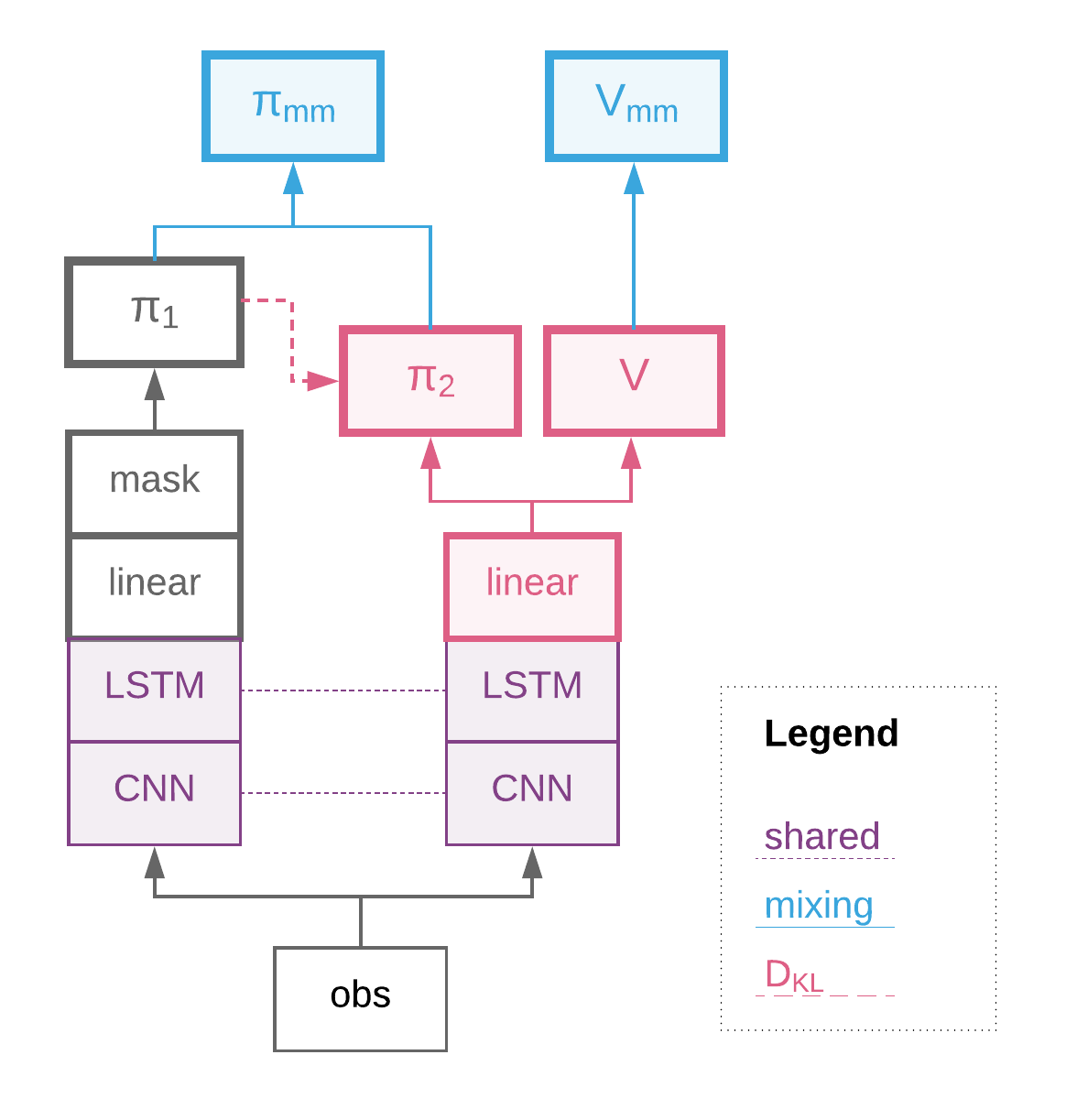} 
\caption{\methodshort for action spaces progression}
\label{fig:mm_scheme_as}
\end{subfigure}
\begin{subfigure}[b]{0.3\textwidth}
\centering
\includegraphics[height=5cm]{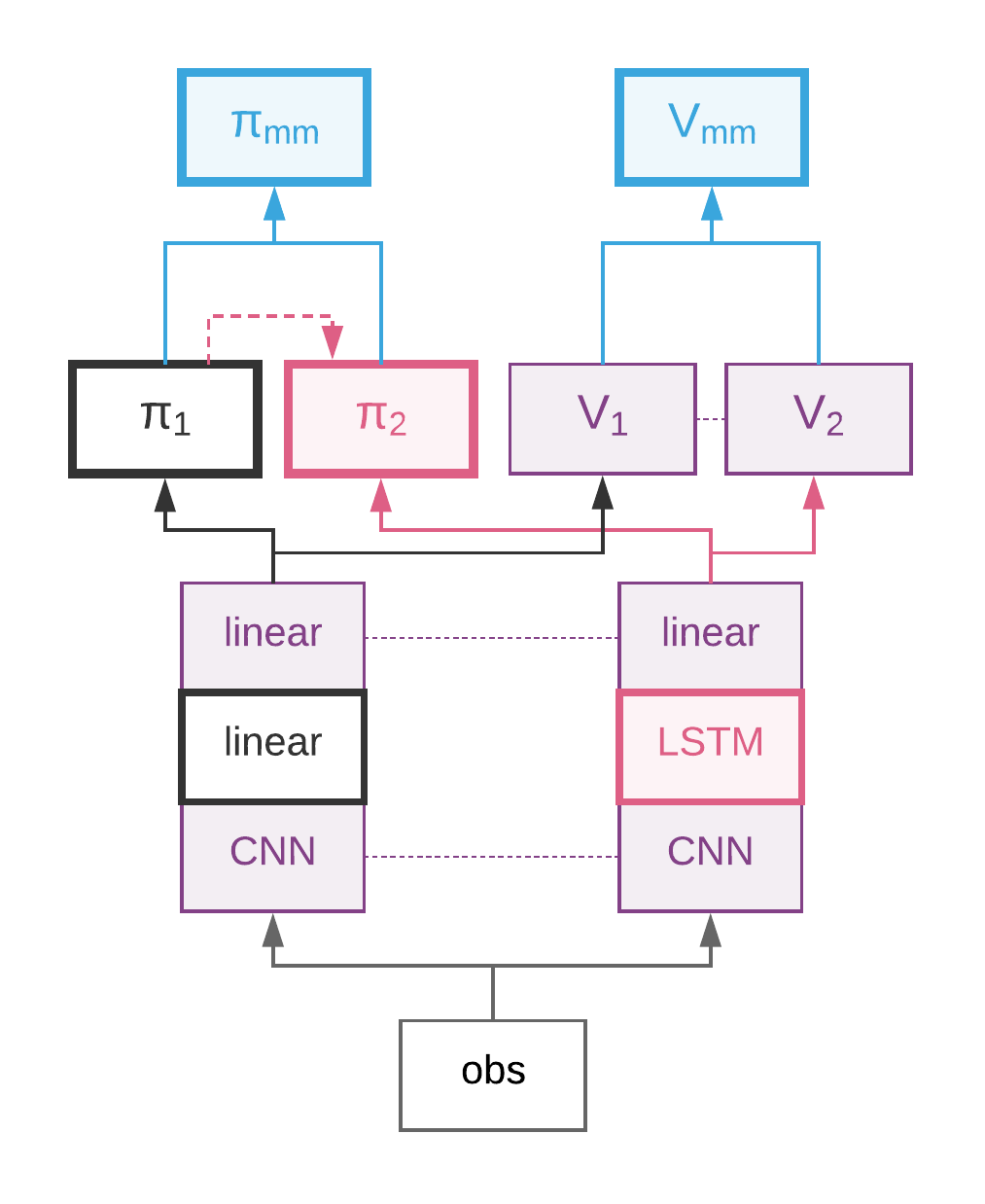} 
\caption{\methodshort for architecture progression}
\label{fig:mm_scheme_core}
\end{subfigure}
\begin{subfigure}[b]{0.3\textwidth}
\centering
\includegraphics[height=5cm]{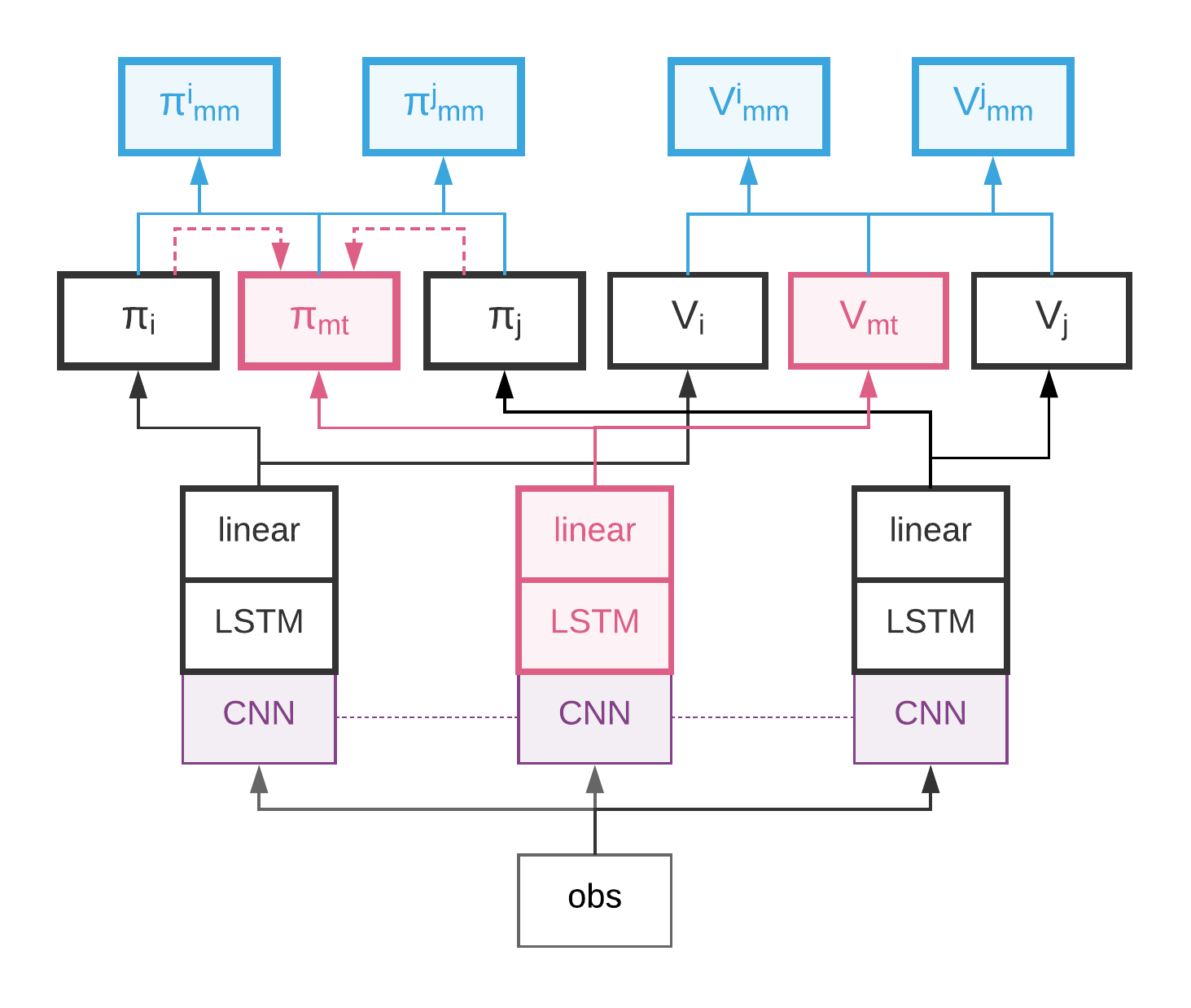} 
\caption{\methodshort for multitask progression}
\label{fig:mm_scheme_mt}
\end{subfigure}
\caption{
Schemes of variours settings \methodshort can be applied to, explored in this paper. Violet nodes represent modules that are shared between $\pi_i$, grey ones are separate modules for the helper agents and red ones -- modules which are unique to the final agent. Blue nodes are the ones that are exposed to the environment -- control policy(ies) and value function(s).
}
\end{figure*}

\subsection{Adjusting $\alpha$ through training}

An important component of the proposed method is how to set values of $\alpha$ through time. 
For simplicity let us again consider the case of $K=2$, where one needs just a single $\alpha$ (as $c$ now comes from Bernoulli distribution) which we treat as a function of time $t$.

\paragraph{Hand crafted schedule} Probably the most common approach is to define a schedule by hand.
Unfortunately, this requires per problem fitting, which might be time consuming. Furthermore while designing an annealing schedule is simple (given that we provide enough flat regions so that RL training is stable), following this path might miss the opportunity to learn a better policy using non-monotonic switches in $\alpha$.

\paragraph{Online hyperparameter tuning} 
Since $\alpha$ changes through time one cannot use typical hyperparameter tuning techniques (like grid search or simple Bayesian optimisation) as the space of possible values is exponential in number of timesteps ($\alpha = (\alpha^{(1)}, \cdots, \alpha^{(T)}) \in \triangle^T_{K-1}$, where $\triangle_k$ denotes a $k$ dimensional simplex).
One possible technique to achieve this goal is the recently proposed Population Based Training~\cite{pbt} (PBT) which keeps a population of agents, trained in parallel, in order to optimise hyperparameters through time (without the need of ever reinitialising networks). For the rest of the paper we rely on using PBT for $\alpha$ adaptation, and discuss it in more detail in the next section.

\subsection{Population based training and \methodshort}

Population based training (PBT) is a recently proposed learning scheme, which performs online adaptation of hyperparameters in conjunction with parameter optimisation and a form of online model selection.
As opposed to many classical hyperparameter optimisation schemes-- the ability of of PBT to modify hyperparameters throughout a single training run makes it is possible to discover powerful adaptive strategies \emph{e.g.} auto-tuned learning rate annealing schedules.

The core idea is to train a population of agents in parallel, which periodically query each other to check how well they are doing relative to others. Badly performing agents copy the weights (neural network parameters) of stronger agents and perform local modifications of their hyperparameters. This way poorly performing agents are used to explore the hyperparameters space. 

From a technical perspective, one needs to define two functions -- \texttt{eval} which measures how strong a current agent is and \texttt{explore} which defines how to perturb the hyperparameters. As a result of such runs we obtain agents maximising the \texttt{eval} function. Note that when we refer to an \emph{agent} in the PBT context we actually mean the \methodshort agent, which is already a mixture of constituent agents.

We propose to use one of the two schemes, depending on the characteristics of the problem we are interested in. If the models considered have a clear benefit (in terms of performance) of switching from simple to the more complex model, then all one needs to do is provide \texttt{eval} with performance (\emph{i.e.} reward over $k$ episodes) of the mixed policy. For an \texttt{explore} function for $\alpha$ we randomly add or subtract a fixed value (truncating between 0 and 1). Thus, once there is a significant benefit of switching to more complex one -- PBT will do it automatically. On the other hand, often we want to switch from an unconstrained architecture to some specific, heavily constrained one (where there may not be an obvious benefit in performance from switching). In such setting,  as is the case when training a multitask policy from constituent single-task policies, we can make \texttt{eval} an independent evaluation job which only looks at performance of an agent with $\alpha_K=1$. This way we directly optimise for the final performance of the model of interest, but at the cost of additional evaluations needed for PBT.

\section{Experiments}

We now test and analyse our method on three sets of RL experiments. 
We train all agents with a form of batched actor critic with an off policy correction 
 \cite{impala} using DeepMind Lab~\cite{dmlab} as an environment suite. 
This environment offers a range of challenging 3D, first-person view based tasks (see, appendix) for RL agents. Agents perceive 96 $\times$ 72 pixel based RGB observations and can move, rotate, jump and tag built-in bots.

We start by demonstrating how \methodshort can be used to scale to a large and complex action space. We follow this with results of scaling complexities of the agent architecture and finally on a problem of learning a multitask policy.
In all following sections we do not force $\alpha$ to approach 1, instead we initialise it around $0$ and analyse its adaptation through time.
Unless otherwise stated, the \texttt{eval} function returns averaged rewards from last 30 episodes of the control policy. 
Note, that even though in the experimental sections we use $K=2$, the actual curriculum goes through potentially infinitely many agents being a result of mixing between $\pi_1$ and $\pi_2$. 
Further technical details and descriptions of all tasks are provided in Appendix.

\subsection{Curricula over number of actions used}

\begin{figure*}[t] \centering
\begin{tabular}{cccc}
\includegraphics[width=0.23\textwidth]{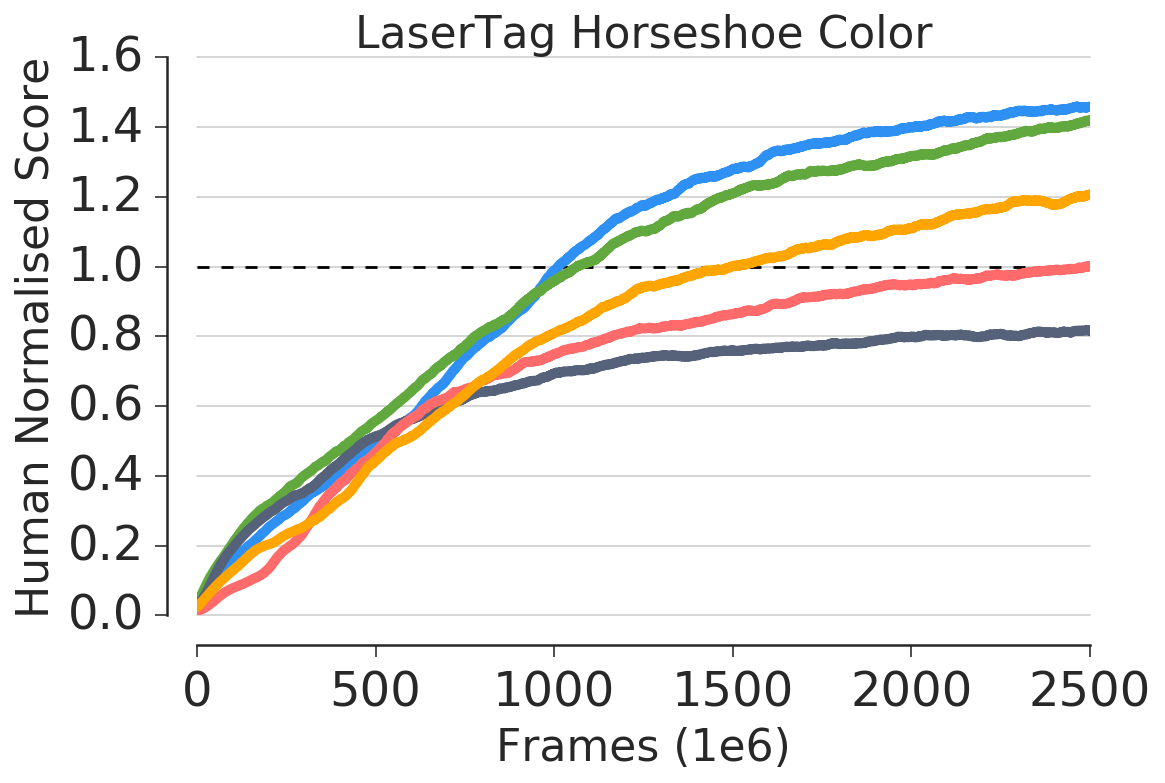} &
\includegraphics[width=0.23\textwidth]{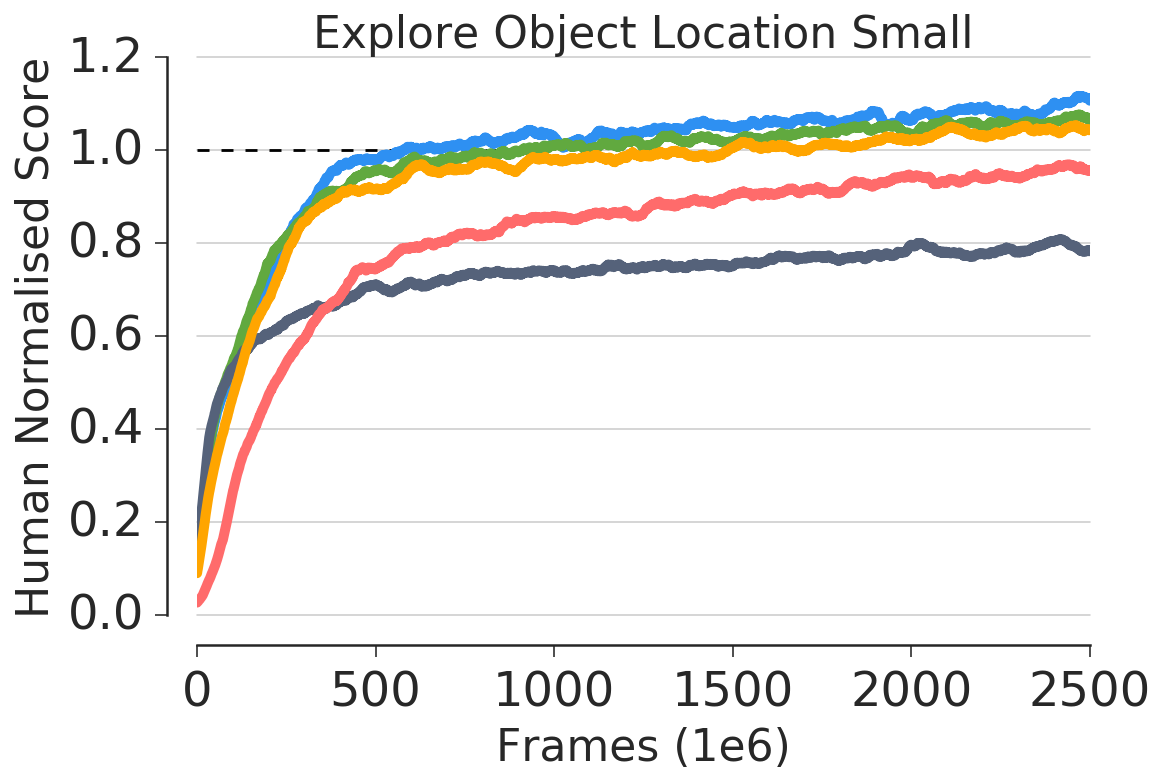} &
\includegraphics[width=0.23\textwidth]{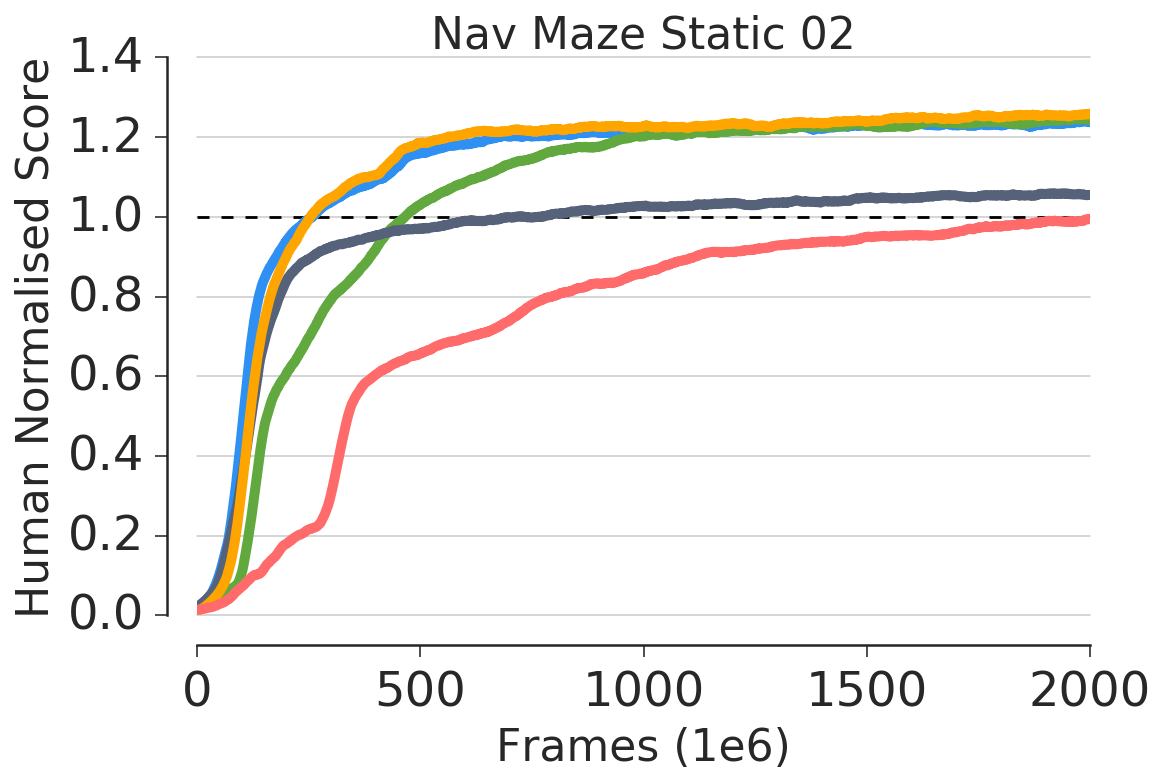} &
\includegraphics[width=0.23\textwidth]{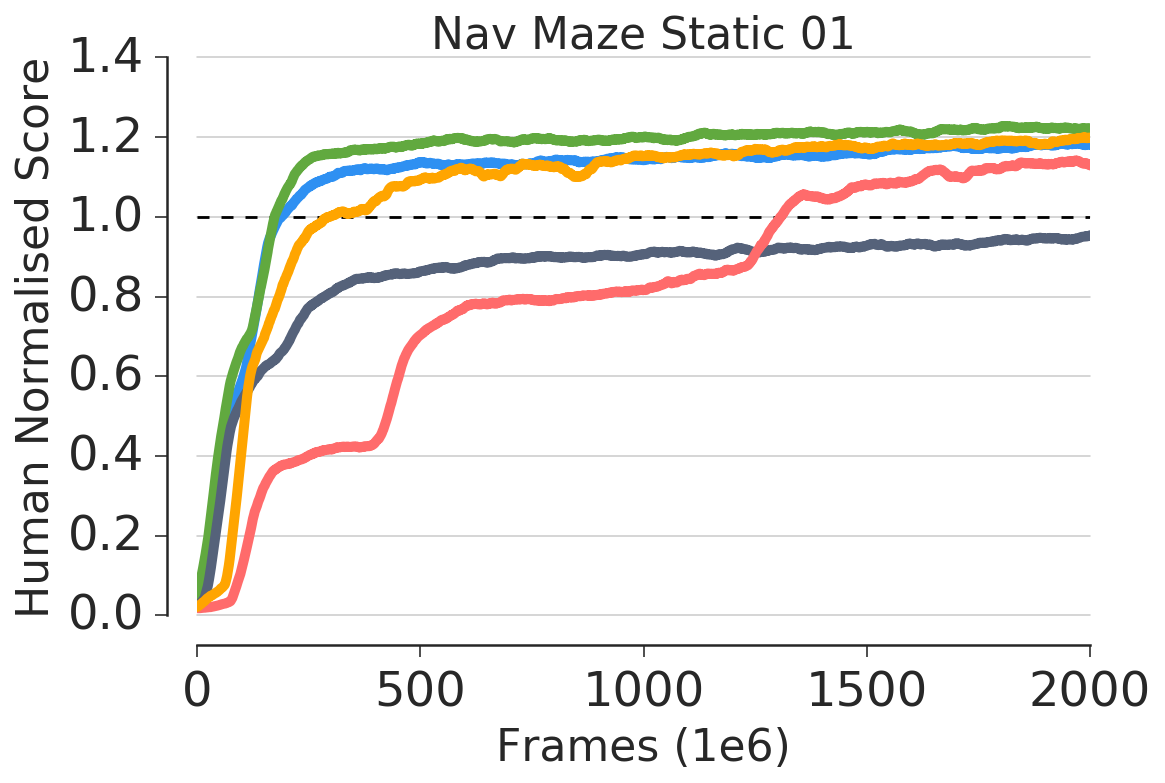} 
\end{tabular}
\centering \includegraphics[width=0.9\textwidth]{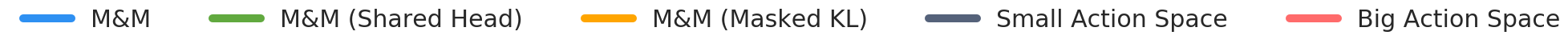}
\caption{Comparison of \methodshort and its variations to baselines applied to the problem of scaling to complex action spaces. Each figure shows the results for a single DMLab level. Each curve shows the mean over 3 populations, each consisting of 10 agents each, with random seeds and hyperparameters as described in the appendix. \methodshort represents the formulation with mixing and a KL distillation cost. The masked KL version only incorporates a KL cost on those actions present in both policies, while the shared head variant shares weights for these common actions directly. All variants of \methodshort outperform the baselines in data efficiency and performance.
}
\label{fig:mm_as}
\end{figure*}

DeepMind Lab provides the agent with a complex action space, represented as a 6 dimensional vector. Two of these action groups are very high resolution (rotation and looking up/down actions), allowing up to 1025 values. The remaining four groups are low resolution actions, such as the ternary action of moving forward, backward or not moving at all, shooting or not shooting etc. If naively approached this leads to around $4 \cdot 10^{13}$ possible actions at each timestep.

Even though this action space is defined by the environment, practitioners usually use an extremely reduced subset of available actions~\cite{mnih2016asynchronous, impala, unreal, mirowski2016learning} -- from 9 to 23 preselected ones. When referring to action spaces we mean the subset of possible actions used for which the agent's policy provides a non zero probability.
  Smaller action spaces significantly simplify the exploration problem and introduce a strong inductive bias into the action space definition. However, having such a tiny subset of possible movements can be harmful for the final performance of the agent. Consequently, we apply \methodshort to this problem of scaling action spaces. We use  9 actions to construct $\pi_1$, the simple policy (called \emph{Small action space}). This is only used to guide learning of our final agent -- which in this case uses 756 actions -- these are all possible combinations of available actions in the environment (when limiting the agent to 5 values of rotation about the z-axis, and 3 values about the x-axis). Similarly to the research in continuous control using diagonal Gaussian distributions~\cite{heess2017emergence} we use a factorised policy (and thus assume conditional independence given state) to represent the joint distribution
$
\pi_2(a_1,a_2,...,a_6|s) := \prod_{j=1}^6 \hat \pi_j( a_j|s ),
$
which we refer to as \emph{Big action space}.
In order to be able to mix these two policies we map $\pi_1$ actions onto the corresponding ones in the action space of $\pi_2$ (which is a strict superset of $\pi_1$).

We use a simple architecture of a convolutional network followed by an LSTM, analogous to previous works in this domain~\cite{unreal}.
For \methodshort we share all elements of two agents apart from the final linear transformation into the policy/value functions (Fig.~\ref{fig:mm_scheme_as}). Full details of the experimental hyper-parameters can be found in the appendix, and on each figure we show the average over 3 runs for each result.

\begin{figure}[h] \centering
\includegraphics[width=0.45\textwidth]{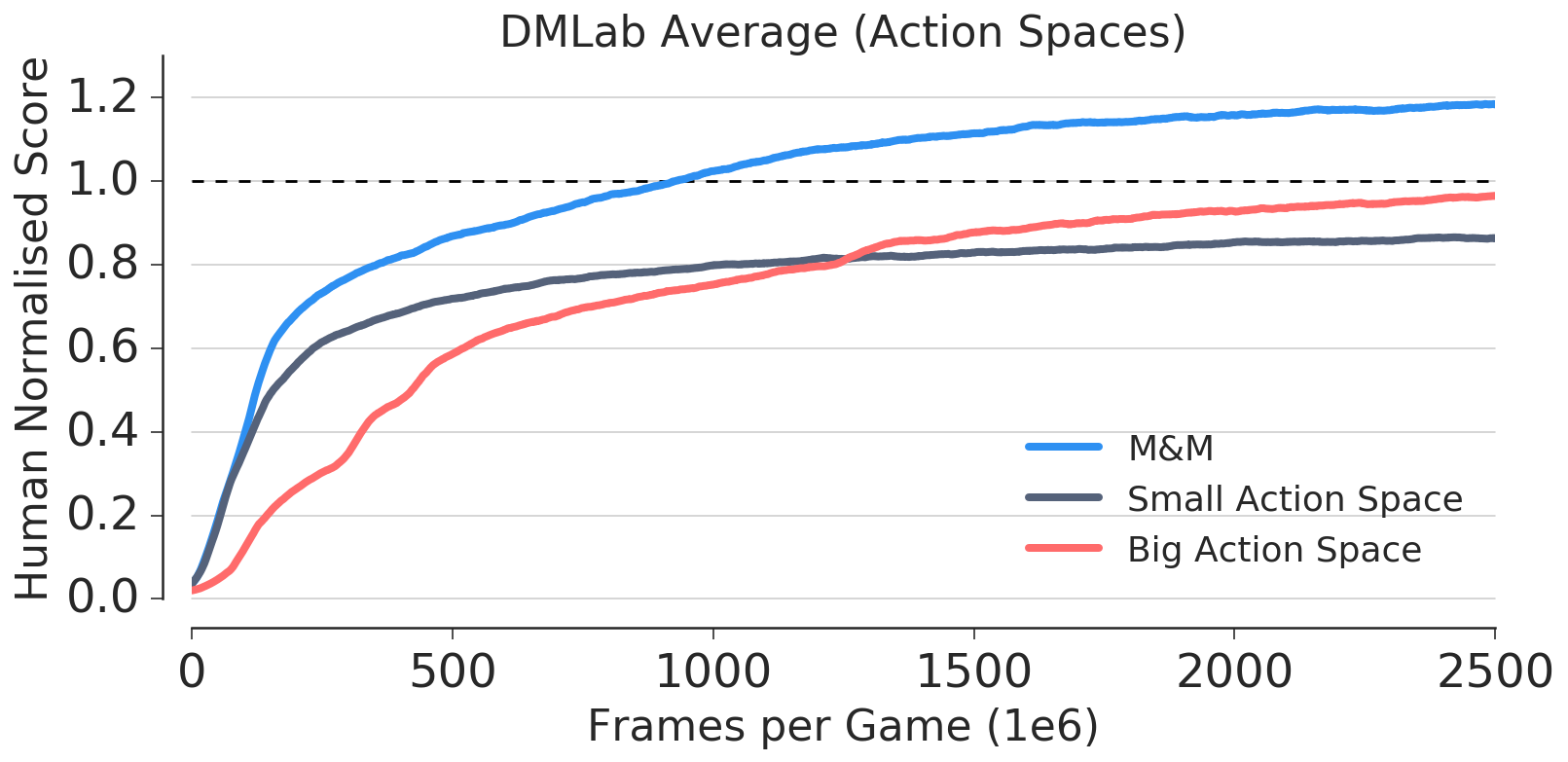} 

\caption{
Human normalised score through training in \textbf{action spaces} experiments. Averaged across levels (for per level curves see Fig.~\ref{fig:mm_as}).
}
\label{fig:mm_as_avg}
\end{figure}

We see that the small action space leads to faster learning but hampers final performance as compared to the big action space  (Fig.~\ref{fig:mm_as} and Fig.~\ref{fig:mm_as_avg}). \methodname applied to this setting gets the best of both worlds -- it learns fast, and not only matches, but surpasses the final performance of the big action space. One possible explanation for this increase is the better exploration afforded by the small action space early on, which allows agents to  exploit fully their flexibility of movement. 

We further compare two variants of our method. We first investigate using  \emph{\methodshort (Shared Head)} -- in this approach, we share weights in the final layer for those actions that are common to both policies. This is achieved by masking the factorised policy $\pi_2$ and renormalising accordingly. We further consider a variant of our distillation cost -- when computing the KL between $\pi_1$ and $\pi_2$ one can also mask this loss such that $\pi_2$ is not penalised for assigning non-zero probabilities to the actions outside  the $\pi_1$ support -- \emph{\methodshort (Masked KL)}. Consistently across tested levels, both shared Head and Masked KL approaches achieve comparable or worse performance than the original formulation. It is worth noting however, that if \methodshort were to be applied to a non-factorised complex action space, the Masked KL might prove beneficial, as it would then be the only signal ensuring agent explore the new actions.

\begin{figure}[h] \centering
\includegraphics[width=4.2cm]{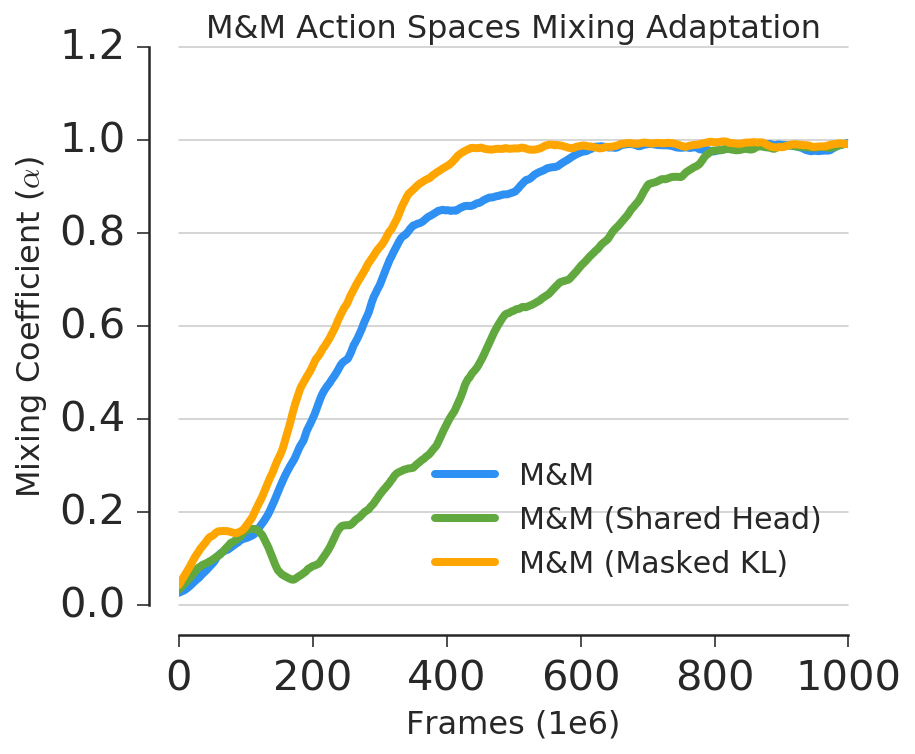}  
\includegraphics[height=3.6cm]{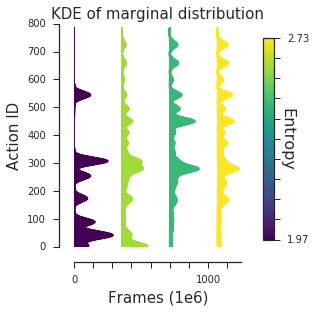} 
\caption{\emph{Left:} Exemplary $\alpha$ value through time from a single run of an action space experiment. \emph{Right:} Progression of marginal distribution of actions taken by the agent. Notice how the collision entropy ($- \mathbb{E} [ \log \sum_a \pi_\methodsymbol^2(a|s)]$) grows over time.} 
\label{fig:mm_as_alpha}
\end{figure}

\begin{figure*}[htb] \centering
\begin{tabular}{cccc}
\includegraphics[width=0.23\textwidth]{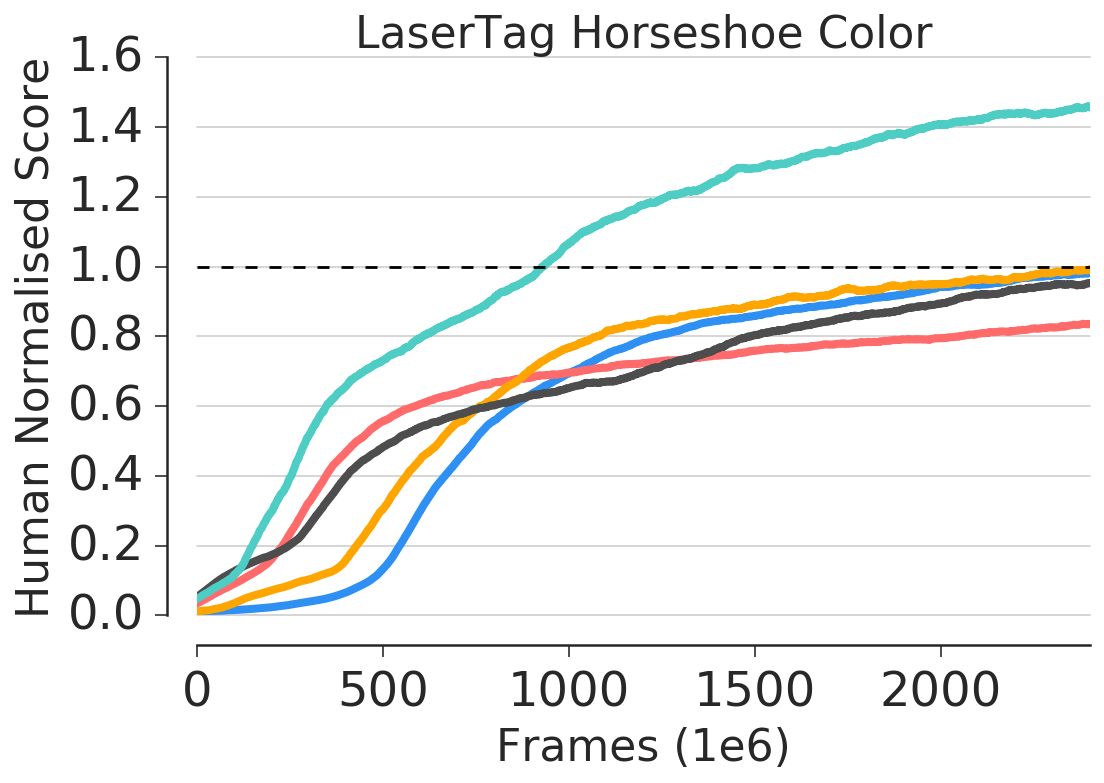} &
\includegraphics[width=0.23\textwidth]{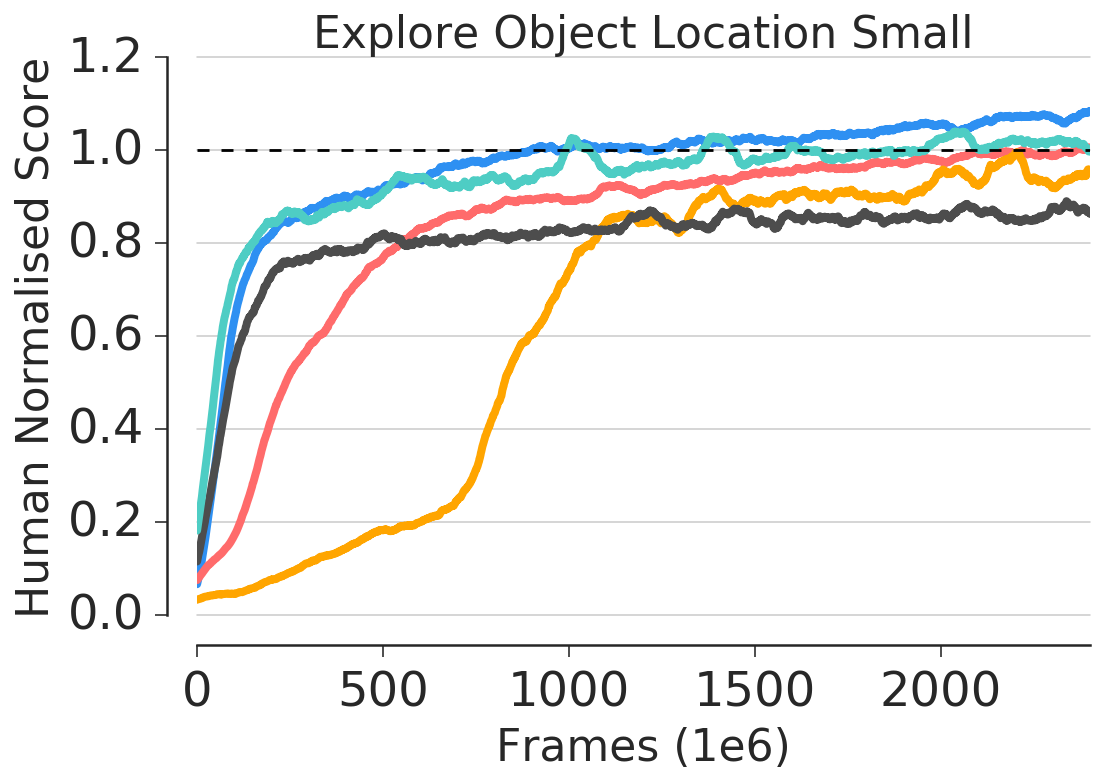} & 
\includegraphics[width=0.23\textwidth]{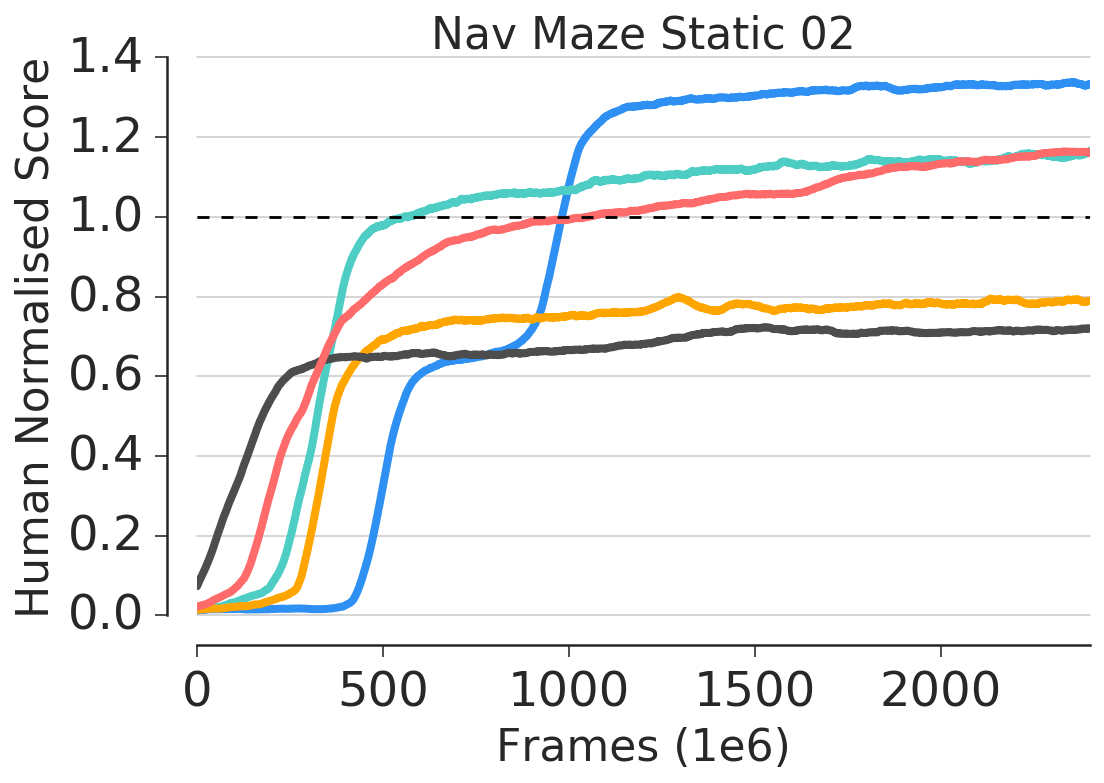} &
\includegraphics[width=0.23\textwidth]{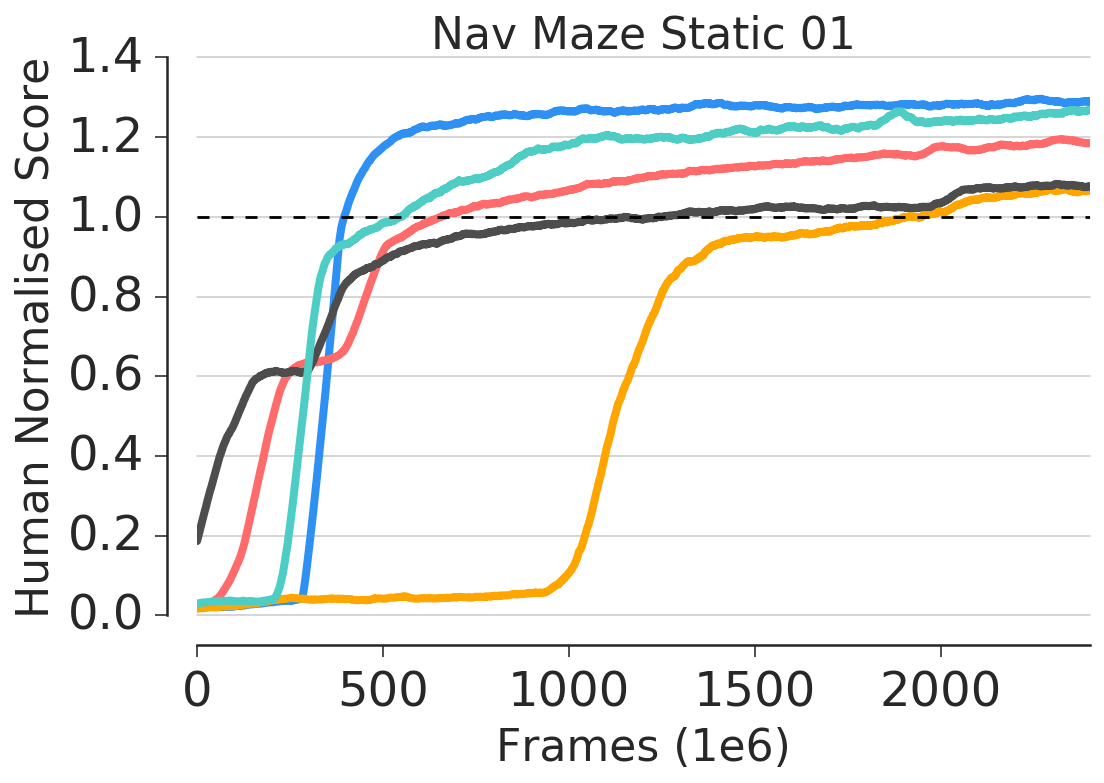} 
\end{tabular}
\centering \includegraphics[width=0.9\textwidth]{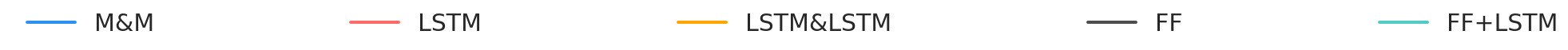}
\caption{Comparison of the \methodshort agent and various baselines on four DM Lab 
levels. Each curve represents the average of 3 independent runs of 10 agents each (used for population based training). FF and LSTM represent the feedforward and LSTM baselines respectively, while FF+LSTM is a model with both cores and a skip connection. FF+LSTM is thus a significantly bigger model than the others, possibly explaining the outlier on the LT level. The LSTM\&LSTM experiment shows \methodshort applied with two LSTM agents. 
}
\label{fig:mm_core}
\end{figure*}
When plotting $\alpha$ through time (Fig.~\ref{fig:mm_as_alpha} \emph{Left}) we see that the agent switches fully to the big action space early on, thus showing that small action space was useful only for initial phase of learning. This is further confirmed by looking at how varied the actions taken by the agent are through training. Fig.~\ref{fig:mm_as_alpha} (\emph{Right}) shows how the marginal distribution over actions evolves through time. We see that new actions are unlocked through training, and further that the final distribution is more entropic that the initial one.

\subsection{Curricula over agent architecture}

Another possible curriculum is over the main computational core of the agent. We use an  architecture analogous to the one used in previous sections, but for the simple or initial agent, we substitute the LSTM with a  linear projection from the processed convolutional signal onto a 256 dimensional latent space. We share both the convolutional modules as well as the policy/value function projections (Fig.~\ref{fig:mm_scheme_core}). We use a 540 element action space, and a factorised policy as described in the previous section.

We ran experiments on four problems in the DM Lab environment, focusing on various navigation tasks. On one hand, reactive policies (which can be represented solely by a FF policy) should learn reasonably quickly to move around and explore, while on the other hand, recurrent networks (which have memory) are needed to maximise the final performance -- by either learning to navigate new maze layouts (Explore Object Location Small) or avoiding (seeking) explored unsuccessful (successful) paths through the maze.

\begin{figure}[h] \centering
\includegraphics[width=0.45\textwidth]{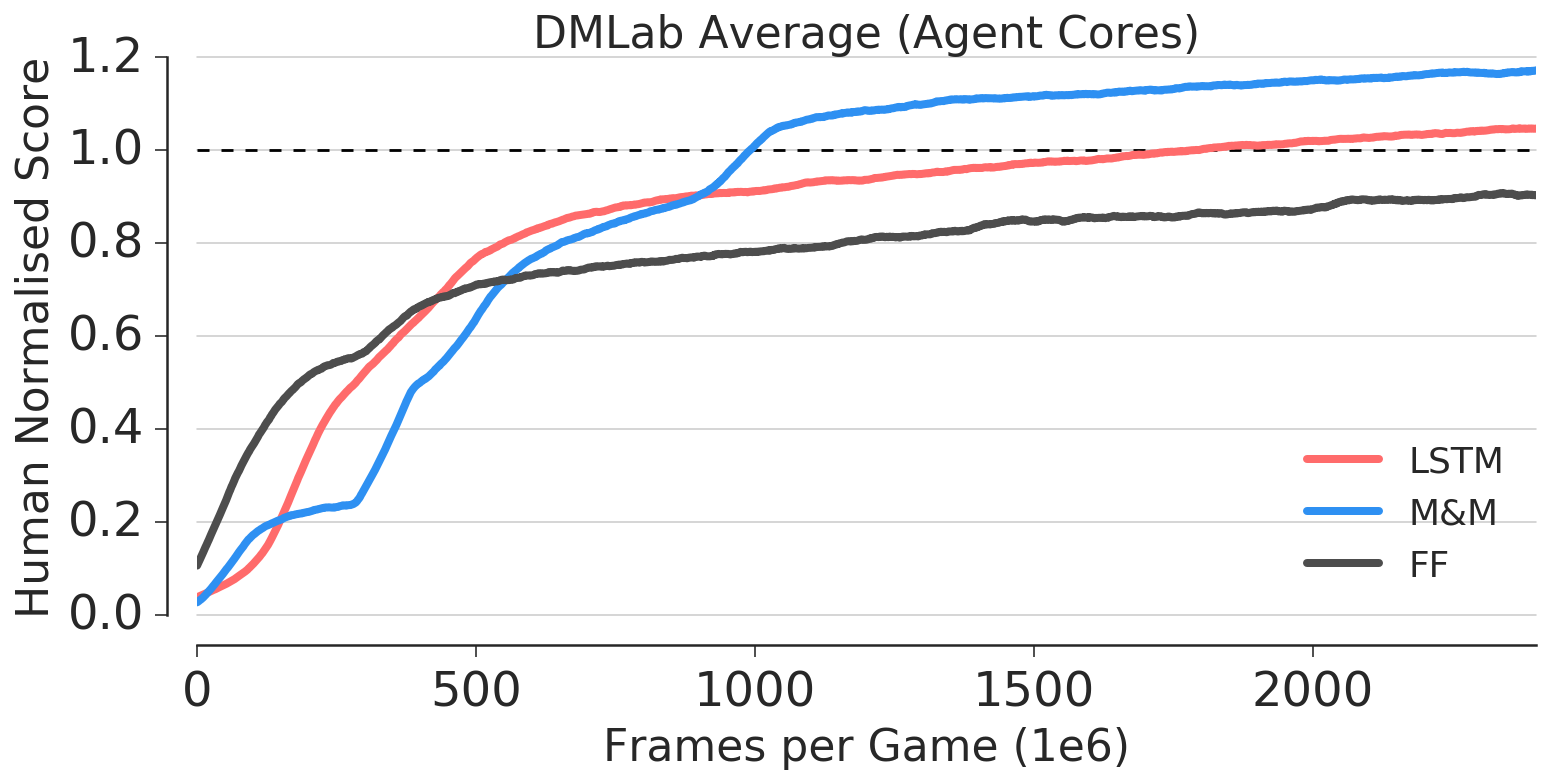} 
\caption{
Human normalised score through training in \textbf{agent's core} experiments. Averaged across levels (for per level curves see Fig.~\ref{fig:mm_core}).
}
\label{fig:mm_core_all}
\end{figure}

As one can see on the average human normalised performance plot (Fig.~\ref{fig:mm_core_all}) the \methodshort applied to the transition between FF and LSTM cores does lead to a significant improvement in final performance (20\% increase in human normalised performance over tasks of interest). It is, however no longer as fast as the FF counterpart. In order to investigate this phenomenon we ran multiple ablation experiments (Fig.~\ref{fig:mm_core}). In the first one, denoted FF+LSTM we use a skip connection which simply adds the activations of the FF core and LSTM core before passing it to a single linear projector for the policy/value heads. This enriched architecture does improve performance of LSTM only model, however it usually learns even slower, and has  very similar learning dynamics to \methodshort.
Consquently it strongly suggests that \methodshort{}'s lack of initial speedup comes from the fact that it is architecturally more similar to the skip connection architecture.
Note, that FF+LSTM is however a significantly bigger model (which appears to be helpful on LT Horseshoe color task). 

Another question of interest is whether the benefit truly comes from the two core types, or  simply through some sort of regularisation effect introduced by the KL cost. To test this hypothesis we also ran an \methodshort-like model but with 2 LSTM cores (instead of the feedforward). This model significantly underperforms all other baselines in speed and performance. This seems to suggest that the distillation or KL cost on its own is not responsible for any benefits we are seeing, and rather it is the full proposed \methodname method.

\begin{figure}[h] \centering
\includegraphics[width=0.45\textwidth]{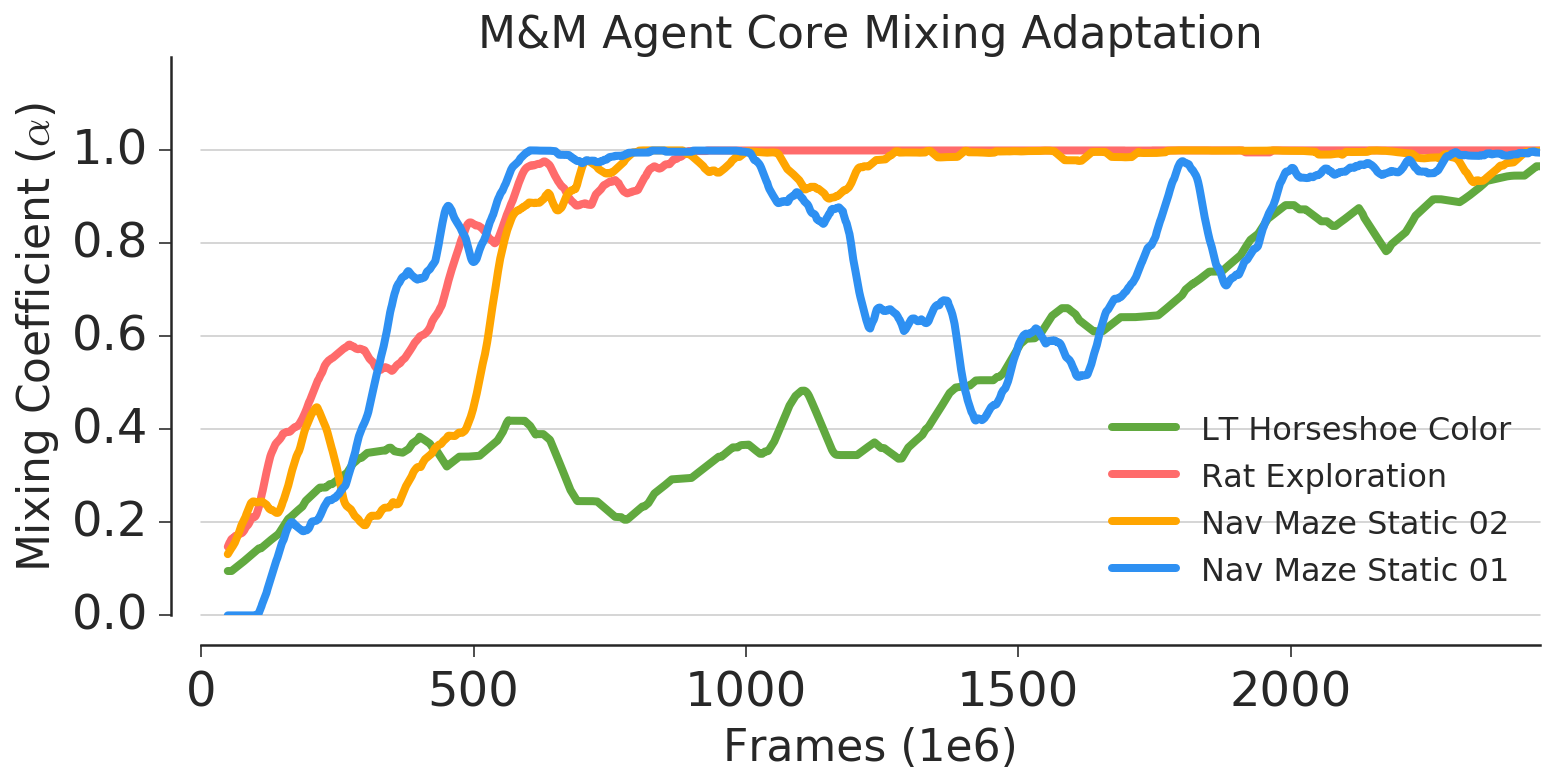} 
\caption{
Exemplary $\alpha$ values through time from single runs of four DMLab levels.
We find that our PBT optimisation finds curricula of different lengths as suited to the target problem. 
}
\label{fig:mm_core_alpha}
\end{figure}

\begin{figure*}[htb] \centering
\begin{tabular}{ccc}
\includegraphics[width=0.3\textwidth]{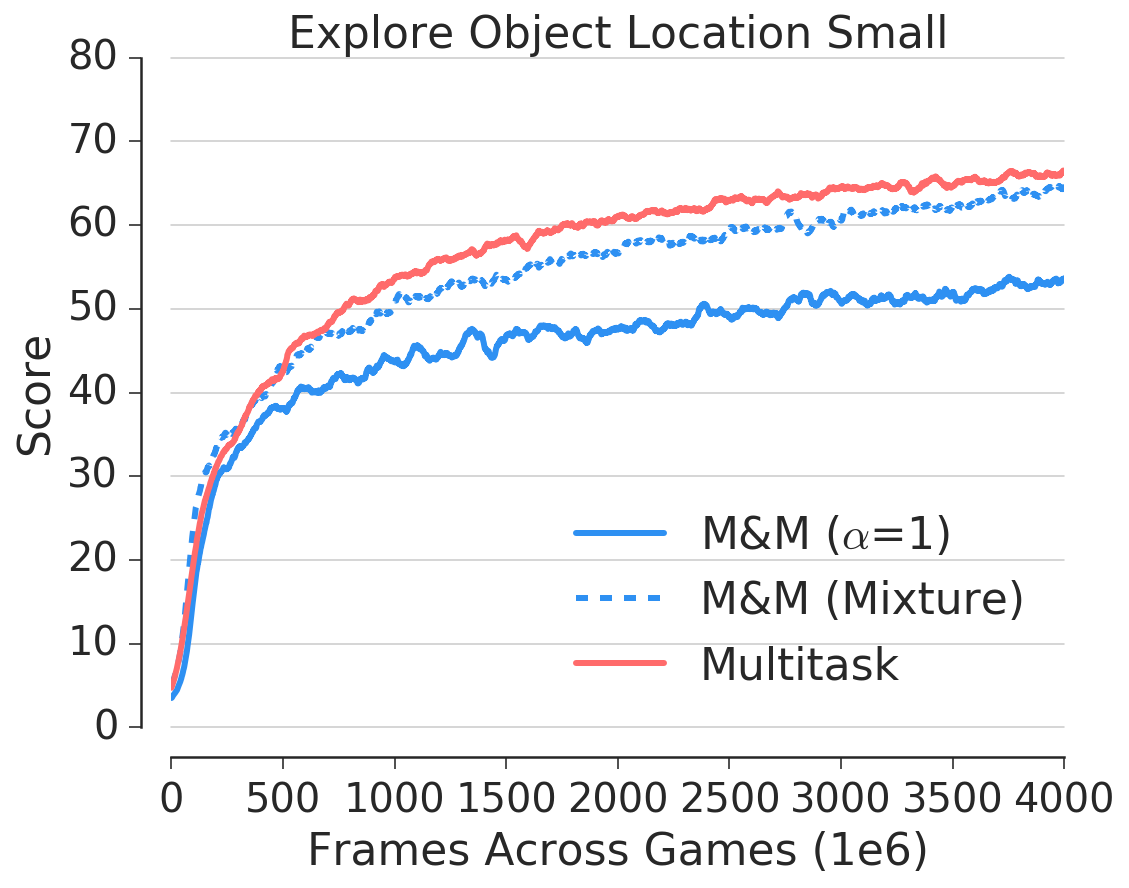} &
\includegraphics[width=0.3\textwidth]{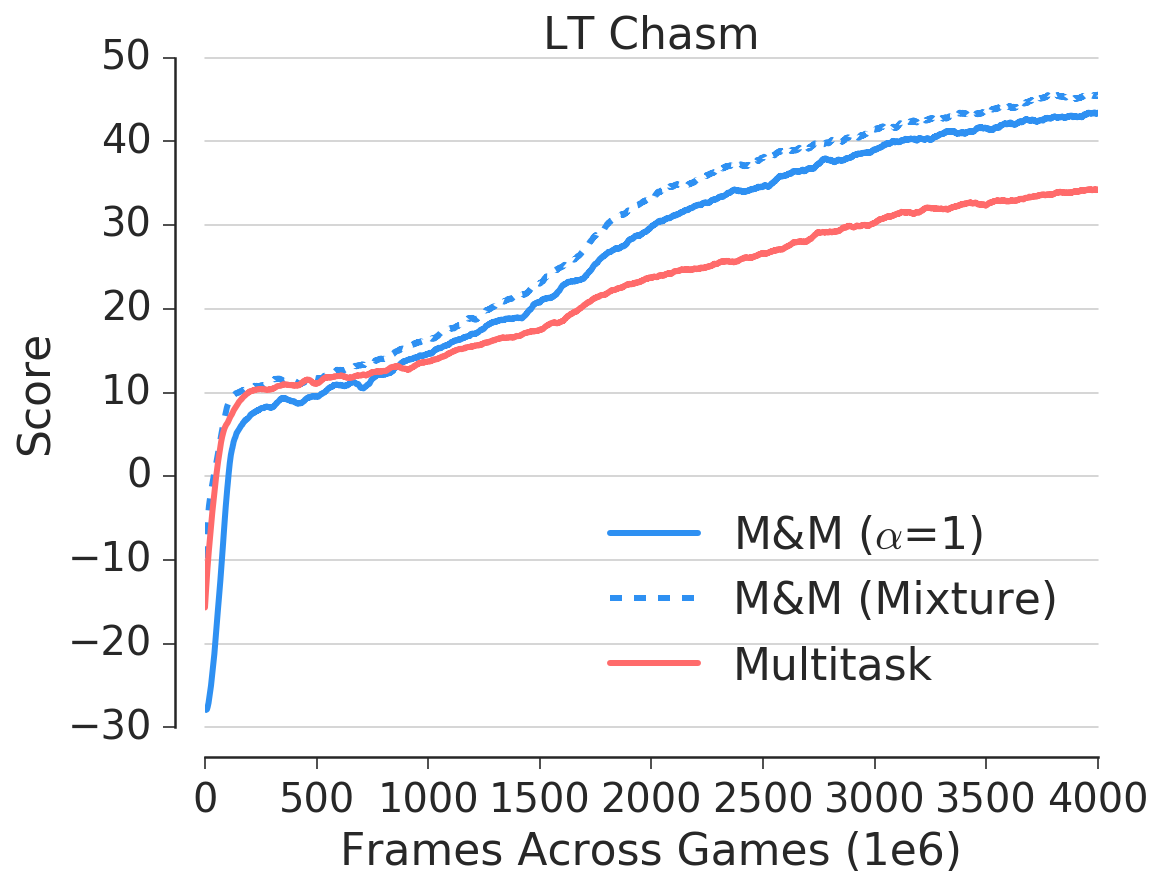} & 
\includegraphics[width=0.3\textwidth]{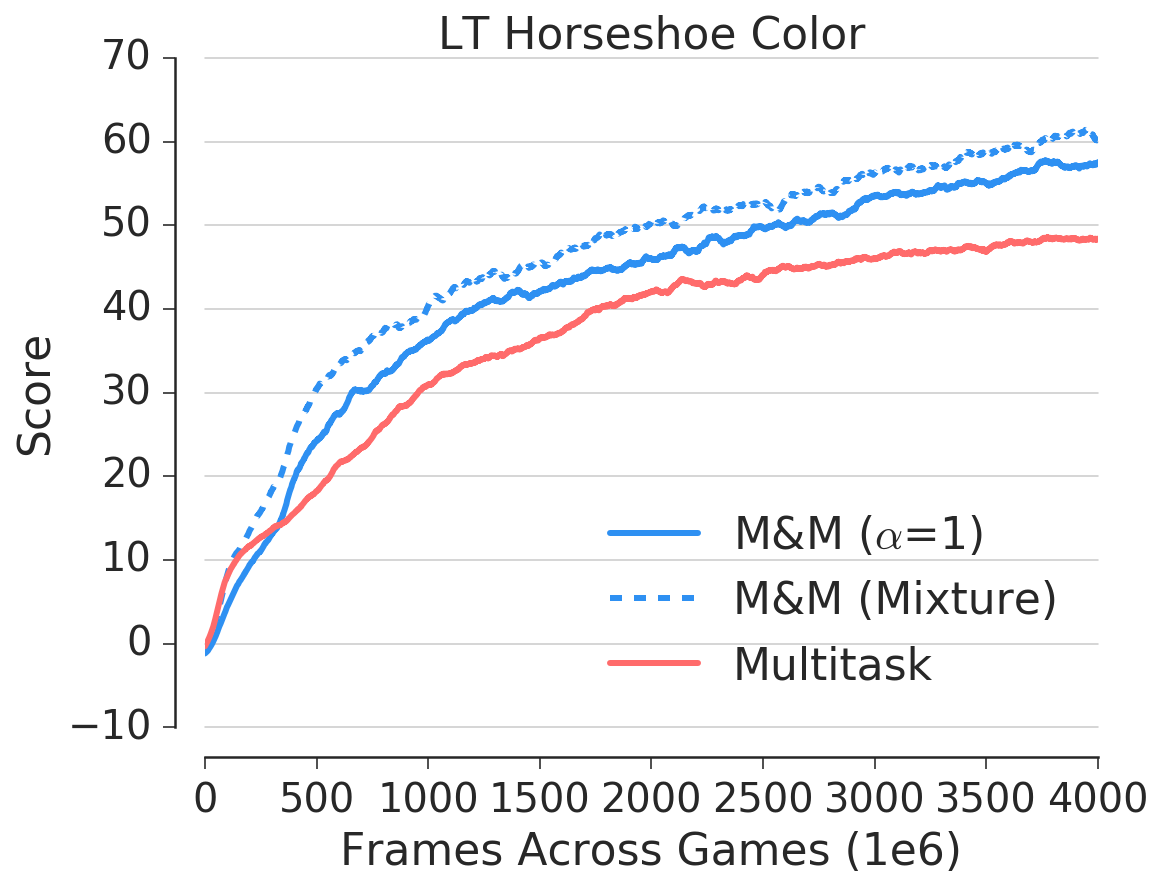} 
\end{tabular}
\centering 
\caption{Performance of \methodshort applied to the multitask domain on three problems considered. Note that this is a performance of an agent trained jointly on all these tasks. The x-axis counts frames across tasks and the y-axis shows score per episode.}
\label{fig:mm_mt}
\end{figure*}

Finally if we look at the progression of the mixing coefficient ($\alpha$) through time (Fig.~\ref{fig:mm_core_alpha}), we notice once again quick switches on navigation-like tasks (all curves except the green one). However, there are two interesting observations to be made. First, the lasertag level, which requires a lot of reactiveness in the policy, takes much longer to switch to the LSTM core (however it does so eventually). This might be related to complexity of the level, which has pickup gadgets as well as many opponents, making memory useful much later in training. Secondly, for the simple goal finding task in a fixed maze (Nav maze static 01, the blue curve) the agent first rapidly switches to the LSTM, but then more or less mid training switches to the mixture policy ($\alpha \approx 0.5$) while finally switch completely again towards the end of training. This particular behaviour is possible due to the use of unconstrained $\alpha$ adaptation with PBT -- thus depending on the current performance the agent can go back and forth through curriculum, which for this particular problem seems to be needed.

\subsection{Curricula for multitask}

 As a final proof of concept we consider the task of learning a single policy capable of solving multiple RL problems at the same time. The basic approach for this sort of task is to train a model in a mixture of environments or equivalently to train a shared model in multiple environments in parallel \cite{distral,impala}. However, this sort of training can suffer from two drawbacks. First, it is heavily reward scale dependent, and will be biased towards high-reward environments. Second, environments that are easy to train provide a lot of updates for the model and consequently can also bias the solution towards themselves.

To demonstrate this issue we use three DeepMind Lab environments -- one is Explore Object Locations Small, which has high rewards and a steep initial learning curve (due to lots of reward signal coming from gathering apples). The two remaining ones are challenging laser tag levels (described in detail in the appendix). In both these problems training is hard, as the agent is interacting with other bots as well as complex mechanics (pick up bonuses, tagging floors, etc.). 

We see in Fig.~\ref{fig:mm_mt} that the multitask solution focuses on solving the navigation task, while performing comparitively poorly on the more challenging problems. To apply \methodshort to this problem we construct one agent per environment (each acting as $\pi_1$ from previous sections) and then one centralised ``multitask'' agent ($\pi_2$ from previous sections). Crucially, agents share convolutional layers but have independent LSTMs. Training is done in a multitask way, but the control policy in each environment is again a mixture between the task specific $\pi_i$ (the specialist) and $\pi_\mathrm{mt}$ (centralised agent), see Fig.~\ref{fig:mm_scheme_mt} for details. Since it is no longer beneficial to switch to the centralised policy, we use the performance of  $\pi_\mathrm{mt}$ (i.e. the central policy) as the optimisation criterion (eval) for PBT, instead of the control policy. 

We evaluate both the performance of the mixture and the centralised agent independently. Fig.~\ref{fig:mm_mt} shows per task performance of the proposed method. One can notice much more uniform performance -- the \methodshort agent learns to play well in both challenging laser tag environments, while slightly sacrificing performance in a single navigation task. One of the reasons of this success is the fact that knowledge transfer is done in policy space, which is invariant to reward scaling.  While the agent can still focus purely on high reward environments once it has switched to using only the central policy, this inductive bias in training with \methodshort ensures a much higher minimum score.

\section{Conclusions}

We have demonstrated that the proposed method {\textendash} {\methodname } {\textendash} is an effective training framework to both improve final performance and accelerate the learning process for complex agents in challenging environments.
This is achieved by constructing an implicit curriculum over agents of different training complexities. The collection of agents is bound together as a single composite whole using a mixture policy. Information can be shared between the components via shared experience or shared architectural elements, and also through a distillation-like KL-matching loss. Over time the component weightings of this mixture are adapted such that at the end of training we are left with a single active component consisting of the most complex agent -- our main agent of interest from the outset.
From an implementation perspective, the proposed method can be seen as a simple wrapper (the \methodshort wrapper) that is compatible with existing agent architectures and training schemes; as such it could easily be introduced as an additional element in conjunction with wide variety of on- or off-policy RL algorithms. In particular we note that, despite our focus on policy-based agents in this paper, the principles behind \methodname are also easily applied to value-based approaches such as Q-learning.

By leveraging \methodshort training, we are able to train complex agents much more effectively and in much less time than is possible if one were to attempt to train such an agent without the support of our methods.
The diverse applications presented in this paper support the generality of our approach.
We believe our training framework could help the community unlock the potential of powerful, but hitherto intractable, agent variants.

\section*{Acknowledgements}

We would like to thank Raia Hadsell, Koray Kavukcuoglu, Lasse Espeholt and Iain Dunning for their invaluable comments, advice and support.

\appendix{}

\section*{Appendix}

 \begin{figure*}[t]
 \centering
 \includegraphics[width=0.33\textwidth]{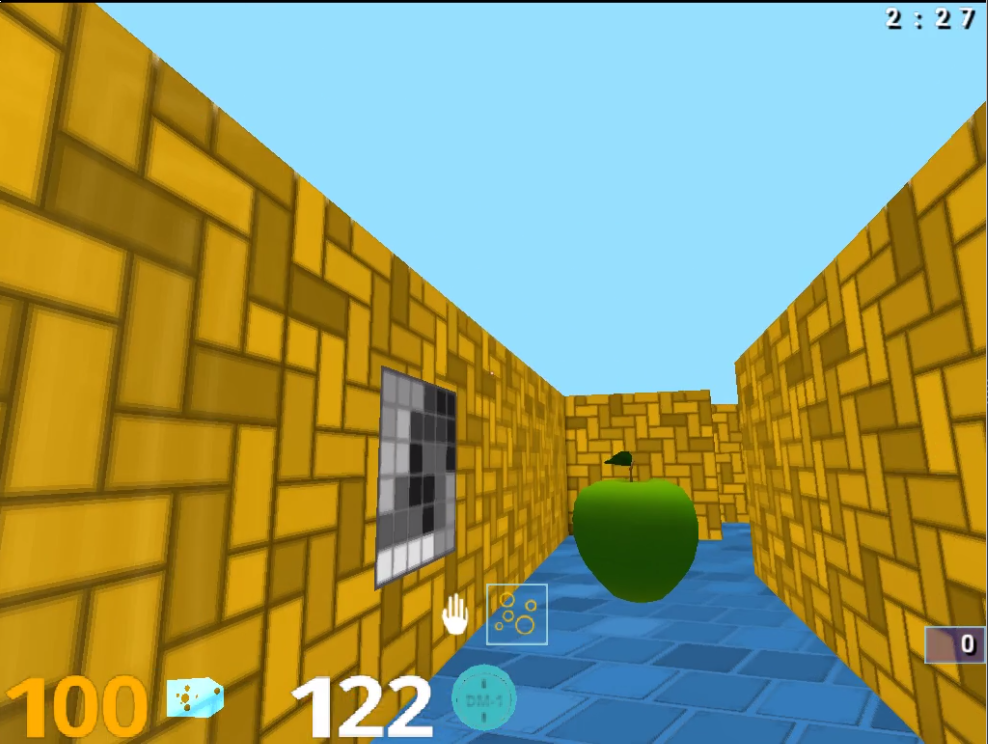}
 \includegraphics[width=0.33\textwidth]{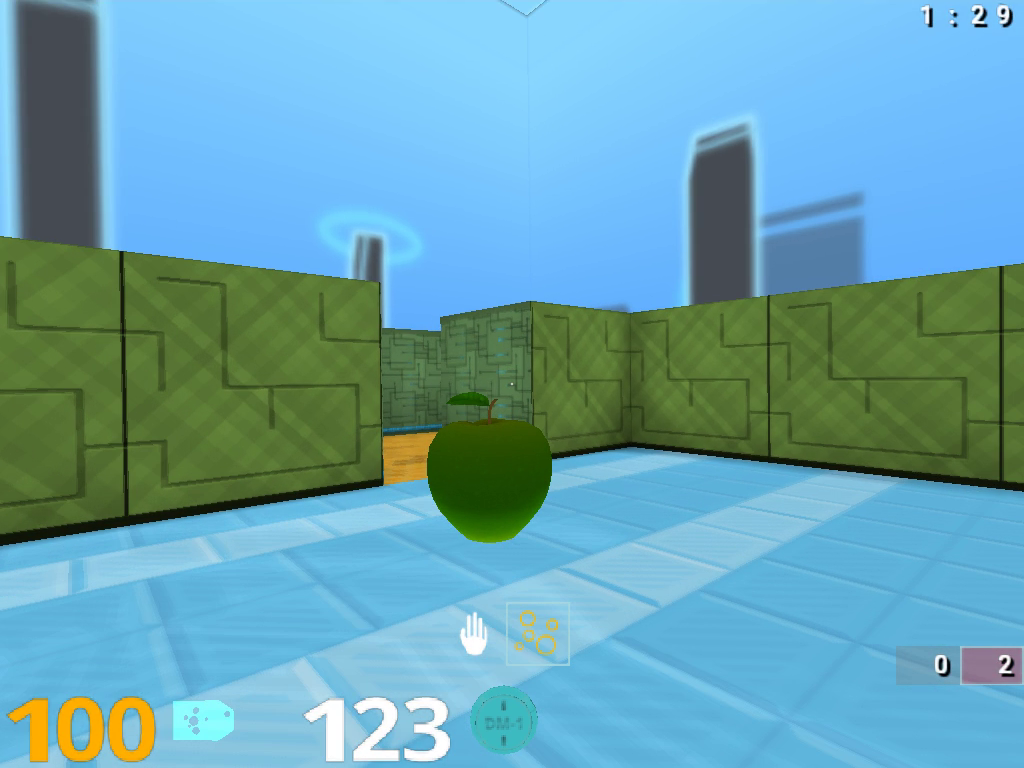}
 \includegraphics[width=0.33\textwidth]{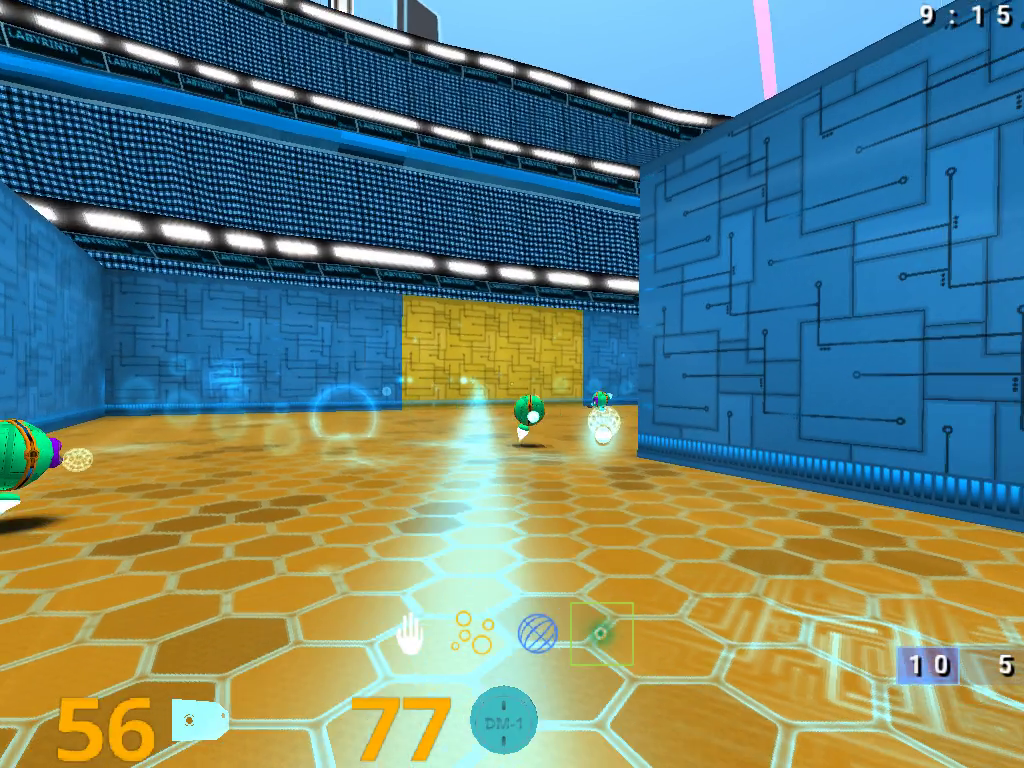}
 \caption{Exemplary tasks of interest from DM Lab environment~\cite{dmlab}. From left:
 \emph{Nav maze static 02} -- The task involves finding apples (+1 reward) and a final goal (+10 reward) in a fixed maze. Every time an agent finds the goal it respawns in a random location, and all objects respawn too.
 \emph{Explore Object Locations Small} -- The task is to navigate through a 3D maze and eat all the apples (each gives +1 reward). Once it is completed, the task restarts. Each new episode differs in terms of apples locations, map layout as well as visual theme. \emph{Lt Horseshoe Color} -- the task is a game of lasertag, where player tries to tag as many of high skilled built-in bots as possible, while using pick-up gadgets (which enhance tagging capabilities). 
 }
 \label{fig:dm_lab}
 \end{figure*}
 
\section{Network architectures}

Default network architecture consists of:
\begin{itemize}
\item Convolutional layer with 16 8x8 kernels of stride 4
\item ReLU
\item Convolutional layer with 32 4x4 kernels of stride 2
\item ReLU
\item Linear layer with 256 neurons
\item ReLU
\item Concatenation with one hot encoded last action and last reward
\item LSTM core with 256 hidden units
    \begin{itemize}
    \item Linear layer projecting onto policy logits, followed by softmax
    \item Linear layer projecting onto baseline
    \end{itemize}
\end{itemize}

Depending on the experiment, some elements are shared and/or replaced as described in the text. 

\section{PBT~\cite{pbt} details}

In all experiments PBT controls adaptation of three hyperparameters: $\alpha$, learning rate and entropy cost regularisation. We use populations of size 10.

The \texttt{explore} operator for learning rate and entropy regularisation is the permutation operator, which randomly multiplies the corresponding value by $1.2$ or $0.8$. For $\alpha$ it is an adder operator, which randomly adds or substracts $0.05$ and truncates result to $[0,1]$ interval. Exploration is executed with probability 25\% independently each time worker is ready.

The \texttt{exploit} operator copies all the weights and hyperparameters from the randomly selected agent if it's performance is significantly better.

Worker is deemed ready to undergo adaptation each 300 episodes. 

We use T-Test with p-value threshold of 5\% to answer the question whether given performance is  significantly better than the other, applied to averaged last 30 episodes returns.

Initial distributions of hyperparameters are as follows:
\begin{itemize}
\item learning rate: loguniform(1e-5, 1e-3)
\item entropy cost: loguniform(1e-4, 1e-2)
\item alpha: loguniform(1e-3, 1e-2)
\end{itemize}

\subsection{Single task experiments}

The \texttt{eval} function uses $\pi_\methodsymbol$ rewards. 

\subsection{multitask experiments}

The \texttt{eval} function uses $\pi_\mathrm{mt}$ rewards, which requires a separate evaluation worker per learner.

\section{\methodshort details}

$\lambda$ for action space experiments is set to $1.0$, and for agent core and multitask to $100.0$.
In all experiments we allow backpropagation through both policies, so that teacher is also regularised towards student (and thus does not diverge too quickly), which is similar to Distral work.

While in principle we could also transfer knowledge between value functions, we did not find it especially helpful empirically, and since it introduces additional weight to be adjusted, we have not used it in the reported experiments. 

\section{IMPALA~\cite{impala} details}

We use 100 CPU actors per one learner. Each learner is trained with a single K80 GPU card.
We use vtrace correction with truncation as described in the original paper.

Agents are trained with a fixed unroll of 100 steps. Optimisation is performned using RMSProp with decay of 0.99, epsilon of 0.1. Discounting factor is set to 0.99, baseline fitting cost is 0.5, rewards are clipped at 1. Action repeat is set to 4.

\section{Environments}

We ran DM Lab using 96 $\times$ 72 $\times$ 3 RGB observations, at 60 fps.

\subsection{Explore Object Locations Small}

The task is to find all apples (each giving 1 point) in the procedurally generated maze, where each episode has different maze, apples locations as well as visual theme. Collecting all apples resets environment.

\subsection{Nav Maze Static 01/02}

Nav Maze Static 01 is a fixed geometry maze with apples (worth 1 point) and one calabash (worth 10 points, getting which resets environment). Agent spawns in random location, but walls, theme and objects positions are held constant.

The only difference for Nav Maze Static 02 is that it is significantly bigger.

\subsection{LaserTag Horseshoe Color}

Laser tag level against 6 built-in bots in a wide horseshoe shaped room. There are 5 Orb Gadgets and 2 Disc Gadgets located in the middle of the room, which can be picked up and used for more efficient tagging of opponents.

\subsection{LaserTag Chasm}

Laser tag level in a square room with Beam Gadgets, Shield Pickups (50 health) and Overshield Pickups (50 armor) hanging above a tagging floor (chasm) splitting room in half. Jumping is required to reach the items. Falling into the chasm causes the agent to lose 1 point. There are 4 built-in bots.

\section{Proofs}
First let us recall the loss of interest
\begin{equation}
\mathcal{L}_\methodsymbol(\theta) = \frac{1-\alpha}{|S|} \sum_{s \in S} \sum_{t=1}^{|s|}  \dkl(\pi_1(\cdot|s_t) \| \pi_2(\cdot|s_t)),
\end{equation}
where each $s \in S$ come from $\pi_\methodsymbol = (1-\alpha)\pi_1 + \alpha \pi_2$.

\textbf{Proposition 1.}\emph{
Lets assume we are given a set of $N$ trajectories from some predefined mix $\pi_\methodsymbol = (1-\alpha)\pi_1 + \alpha \pi_2$ for any fixed $\alpha \in (0,1)$ and a big enough neural network with softmax output layer as $\pi_2$.
Then in the limit as $N \rightarrow \infty$, the minimisation of Eq. 1 converges to $\pi_1$ if the optimiser used is globally convergent when minimising cross entropy over a finite dataset.
}
\begin{proof}
For  $D_N$ denoting set of $N$ sampled trajectories over state space $\mathcal{S}$ let as denote by $\hat{ \mathcal{S} }_N$ the set of all states in $D_N$, meaning that $\hat{ \mathcal{S}}_N = \cup D_N$.
Since $\pi_2$ is a softmax based policy, it assigns non-zero probability to all actions in every state. Consequently also $\pi_\methodsymbol$ does that as $\alpha \in (0,1)$. Thus  we have
$$
\lim_{N \rightarrow \infty} \hat{ \mathcal{S}}_N  =  \mathcal{S}.
$$
Due to following the mixture policy, actual dataset $\hat D_N$ gathered can consist of multiple replicas of each element in $\hat{ \mathcal{S} }_N$, in different proportions that one would achieve when following $\pi_1$.
Note, note however that if we use optimiser which is capable of minimising the cross entropy over finite dataset, it can also minimise loss (1) over $\hat D_N$ thus in particular over $\hat{\mathcal{S}}_N$ which is its strict subset. 
Since the network is big enough, it means that it will converge to 0 training error:
$$
\forall_{s \in \hat{ \mathcal{S}}_N }\lim_{t \rightarrow \infty} \dkl(\pi_1(a|s) \| \pi_2(a|s, \theta_t)) = 0
$$
where $\theta_t$ is the solution of $t$th iteration of the optimiser used. 
Connecting the two above we get that in the limit of $N$ and $t$ 
$$
\forall_{s \in \mathcal{S}} \dkl(\pi_1(a|s_t) \| \pi_2(a|s_t, \theta_t)) = 0 \iff \pi_1 = \pi_2.
$$
\end{proof}
While the global convergence might sound like a very strong property, it holds for example when both teacher and student policies are linear. In general for deep networks it is hypothesised that if they are big enough, and well initialised, they do converge to arbitrarily small training error even if trained with a simple gradient descent, thus the above proposition is not too restrictive for Deep RL.

\section{On $\alpha$ based scaling of knowledge transfer loss}

Let as take a closer look at the proposed loss
$$
\ell_\methodsymbol(\theta) = (1-\alpha) \dkl(\pi_1(\cdot|s) \| \pi_2(\cdot|s)) = $$
$$
= (1-\alpha)\mathrm{H}(\pi_1(\cdot|s) \| \pi_2(\cdot|s)) - (1-\alpha)\mathrm{H}(\pi_1(\cdot|s))
$$
and more specifically at $1-\alpha$ factor. The intuitive justification for this quantity is that it leads to $\dkl$ gradually disappearing as \methodshort agent is switching to the final agent.
However, one can provide another explanation. 
Let us instead  consider divergence between mixed policy and the target policy (which also has the property of being $0$ once agent switches):
$$
\hat \ell_\methodsymbol(\theta) = \dkl(\pi_\methodsymbol(\cdot|s) \| \pi_2(\cdot|s)) =  $$
$$
= \mathrm{H}(\pi_\methodsymbol(\cdot|s) \| \pi_2(\cdot|s)) - \mathrm{H}(\pi_\methodsymbol(\cdot|s))=
$$
$$
=\mathrm{H}((1-\alpha) \pi_1(\cdot|s) + \alpha \pi_2(\cdot|s) \| \pi_2(\cdot|s)) - \mathrm{H}(\pi_\methodsymbol(\cdot|s))
$$
$$
=\mathrm{H}((1-\alpha) \pi_1(\cdot|s) \| \pi_2(\cdot|s)) + \alpha \mathrm{H}(\pi_2(\cdot|s) - \mathrm{H}(\pi_\methodsymbol(\cdot|s))
$$
$$
=(1-\alpha) \mathrm{H}( \pi_1(\cdot|s) \| \pi_2(\cdot|s)) - (\mathrm{H}(\pi_\methodsymbol(\cdot|s)) - \alpha \mathrm{H}(\pi_2(\cdot|s))
$$
One can notice, that there are two factors of both losses, one being a cross entropy between $\pi_1$ and $\pi_2$ and the other being a form of entropy regularisers. Furthermore, these two losses differ only wrt. regularisations:
$$
\ell_\methodsymbol(\theta) - \hat \ell_\methodsymbol(\theta) =
$$
$$
= - (1-\alpha)\mathrm{H}(\pi_1(\cdot|s)) + ( \mathrm{H}(\pi_\methodsymbol(\cdot|s)) - \alpha \mathrm{H}(\pi_2(\cdot|s)  )=
$$
$$
= \mathrm{H}(\pi_\methodsymbol(\cdot|s)) - (\alpha \mathrm{H}(\pi_2(\cdot|s) + (1-\alpha)\mathrm{H}(\pi_1(\cdot|s))) 
$$
but since entropy is concave, this quantitiy is non-negative, meaning that 
$$
\ell_\methodsymbol(\theta) \geq \hat \ell_\methodsymbol(\theta)
$$
therefore
$$
 - (1-\alpha)\mathrm{H}(\pi_\methodsymbol(\cdot|s)) \geq - (\mathrm{H}(\pi_\methodsymbol(\cdot|s)) - \alpha \mathrm{H}(\pi_2(\cdot|s))
$$
Thus the proposed scheme is almost equivalent to minimising KL between mixed policy and $\pi_2$ but simply with more severe regularisation factor (and thus it is the upper bound of the $\hat \ell_\methodsymbol$.

Further research and experiments need to be performed to asses quantitative differences between these costs though. In preliminary experiments we ran, the difference was hard to quantify -- both methods behaved similarly well.

\section{On knowledge transfer loss}

Through this paper we focused on using Kulback-Leibler Divergence for knowledge transfer $\dkl(p\|q) = \mathrm{H}(p,q) - \mathrm{H}(p)$. For many distillation related methods, it is actually equivalent to minimising cross entropy (as $p$ is constant), in \methodshort case the situation is more complex. When both $p$ and $q$ are learning $\dkl$ provides a two-way effect -- from one perspective $q$ is pulled towards $p$ and on the other $p$ is mode seeking towards $q$ while at the same time being pushed towards uniform distribution (entropy maximisation). This has two effects, first, it makes it harder for the teacher to get too ahead of the student (similarly to~\cite{distral, zhang2017deep}); second, additional entropy term makes it expensive to keep using teacher, and so switching is preffered.

Another element which has not been covered in depth in this paper is possibility of deep distillation. Apart from matching policies one could include inner activation matching~\cite{ParisottoBS15}, which could be beneficial for deeper models which do not share modules. Furthermore, for speeding up convergence of distillation one could use Sobolev Training~\cite{czarnecki2017sobolev} and match both policy and its Jacobian matrix. Since policy matching was enough for current experiments, none of these methods has been used in this paper, however for much bigger models and more complex domains it might be the necesity as \methodshort depends on ability to rapidly transfer knowledge between agents.

\bibliography{references}

\begin{thebibliography}{27}
\providecommand{\natexlab}[1]{#1}
\providecommand{\url}[1]{\texttt{#1}}
\expandafter\ifx\csname urlstyle\endcsname\relax
  \providecommand{\doi}[1]{doi: #1}\else
  \providecommand{\doi}{doi: \begingroup \urlstyle{rm}\Url}\fi

\bibitem[Ba \& Caruana(2014)Ba and Caruana]{BaDistill14}
Ba, Jimmy and Caruana, Rich.
\newblock Do deep nets really need to be deep?
\newblock In Ghahramani, Z., Welling, M., Cortes, C., Lawrence, N.~D., and
  Weinberger, K.~Q. (eds.), \emph{Advances in Neural Information Processing
  Systems 27}, pp.\  2654--2662. 2014.

\bibitem[Beattie et~al.(2016)Beattie, Leibo, Teplyashin, Ward, Wainwright,
  K{\"{u}}ttler, Lefrancq, Green, Vald{\'{e}}s, Sadik, Schrittwieser, Anderson,
  York, Cant, Cain, Bolton, Gaffney, King, Hassabis, Legg, and Petersen]{dmlab}
Beattie, Charles, Leibo, Joel~Z., Teplyashin, Denis, Ward, Tom, Wainwright,
  Marcus, K{\"{u}}ttler, Heinrich, Lefrancq, Andrew, Green, Simon,
  Vald{\'{e}}s, V{\'{\i}}ctor, Sadik, Amir, Schrittwieser, Julian, Anderson,
  Keith, York, Sarah, Cant, Max, Cain, Adam, Bolton, Adrian, Gaffney, Stephen,
  King, Helen, Hassabis, Demis, Legg, Shane, and Petersen, Stig.
\newblock Deepmind lab.
\newblock \emph{CoRR}, 2016.

\bibitem[Bengio et~al.(2009)Bengio, Louradour, Collobert, and
  Weston]{bengio2009}
Bengio, Yoshua, Louradour, Jerome, Collobert, Ronan, and Weston, Jason.
\newblock Curriculum learning.
\newblock In \emph{ICML}, 2009.

\bibitem[Brockman et~al.(2016)Brockman, Cheung, Pettersson, Schneider,
  Schulman, Tang, and Zaremba]{brockman2016openai}
Brockman, Greg, Cheung, Vicki, Pettersson, Ludwig, Schneider, Jonas, Schulman,
  John, Tang, Jie, and Zaremba, Wojciech.
\newblock Openai gym.
\newblock \emph{arXiv preprint arXiv:1606.01540}, 2016.

\bibitem[Buciluǎ et~al.(2006)Buciluǎ, Caruana, and
  Niculescu-Mizil]{buciluǎ2006model}
Buciluǎ, Cristian, Caruana, Rich, and Niculescu-Mizil, Alexandru.
\newblock Model compression.
\newblock In \emph{Proceedings of the 12th ACM SIGKDD international conference
  on Knowledge discovery and data mining}, pp.\  535--541. ACM, 2006.

\bibitem[Chen et~al.(2016)Chen, Goodfellow, and Shlens]{Chen2015Net2NetAL}
Chen, Tianqi, Goodfellow, Ian~J., and Shlens, Jonathon.
\newblock Net2net: Accelerating learning via knowledge transfer.
\newblock \emph{ICLR}, abs/1511.05641, 2016.

\bibitem[Czarnecki et~al.(2017)Czarnecki, Osindero, Jaderberg, Swirszcz, and
  Pascanu]{czarnecki2017sobolev}
Czarnecki, Wojciech~M, Osindero, Simon, Jaderberg, Max, Swirszcz, Grzegorz, and
  Pascanu, Razvan.
\newblock Sobolev training for neural networks.
\newblock In \emph{Advances in Neural Information Processing Systems}, pp.\
  4281--4290, 2017.

\bibitem[Elman(1993)]{elman93}
Elman, Jeffrey.
\newblock Learning and development in neural networks: The importance of
  starting small.
\newblock In \emph{Cognition}, pp.\  71--99, 1993.

\bibitem[Espeholt et~al.(2018)Espeholt, Soyer, Munos, Simonyan, Mnih, Ward,
  Doron, Firoiu, Harley, Dunning, Legg, and Kavukcuoglu]{impala}
Espeholt, Lasse, Soyer, Hubert, Munos, Remi, Simonyan, Karen, Mnih, Volodymir,
  Ward, Tom, Doron, Yotam, Firoiu, Vlad, Harley, Tim, Dunning, Iain, Legg,
  Shane, and Kavukcuoglu, Koray.
\newblock Impala: Scalable distributed deep-rl with importance weighted
  actor-learner architectures, 2018.

\bibitem[Graves et~al.(2017)Graves, Bellemare, Menick, Munos, and
  Kavukcuoglu]{Graves17}
Graves, Alex, Bellemare, Marc~G., Menick, Jacob, Munos, R{\'{e}}mi, and
  Kavukcuoglu, Koray.
\newblock Automated curriculum learning for neural networks.
\newblock \emph{CoRR}, 2017.

\bibitem[Heess et~al.(2017)Heess, Sriram, Lemmon, Merel, Wayne, Tassa, Erez,
  Wang, Eslami, Riedmiller, et~al.]{heess2017emergence}
Heess, Nicolas, Sriram, Srinivasan, Lemmon, Jay, Merel, Josh, Wayne, Greg,
  Tassa, Yuval, Erez, Tom, Wang, Ziyu, Eslami, Ali, Riedmiller, Martin, et~al.
\newblock Emergence of locomotion behaviours in rich environments.
\newblock \emph{arXiv preprint arXiv:1707.02286}, 2017.

\bibitem[Hinton et~al.(2015)Hinton, Vinyals, and Dean]{hinton2015distilling}
Hinton, Geoffrey, Vinyals, Oriol, and Dean, Jeff.
\newblock Distilling the knowledge in a neural network.
\newblock \emph{arXiv preprint arXiv:1503.02531}, 2015.

\bibitem[Jacobs et~al.(1991)Jacobs, Jordan, Nowlan, and
  Hinton]{jacobs1991adaptive}
Jacobs, Robert~A, Jordan, Michael~I, Nowlan, Steven~J, and Hinton, Geoffrey~E.
\newblock Adaptive mixtures of local experts.
\newblock \emph{Neural computation}, 3\penalty0 (1):\penalty0 79--87, 1991.

\bibitem[Jaderberg et~al.(2017{\natexlab{a}})Jaderberg, Dalibard, Osindero,
  Czarnecki, Donahue, Razavi, Vinyals, Green, Dunning, Simonyan, Fernando, and
  Kavukcuoglu]{pbt}
Jaderberg, Max, Dalibard, Valentin, Osindero, Simon, Czarnecki, Wojciech~M.,
  Donahue, Jeff, Razavi, Ali, Vinyals, Oriol, Green, Tim, Dunning, Iain,
  Simonyan, Karen, Fernando, Chrisantha, and Kavukcuoglu, Koray.
\newblock Population based training of neural networks.
\newblock \emph{CoRR}, 2017{\natexlab{a}}.

\bibitem[Jaderberg et~al.(2017{\natexlab{b}})Jaderberg, Mnih, Czarnecki,
  Schaul, Leibo, Silver, and Kavukcuoglu]{unreal}
Jaderberg, Max, Mnih, Volodymyr, Czarnecki, Wojciech~Marian, Schaul, Tom,
  Leibo, Joel~Z, Silver, David, and Kavukcuoglu, Koray.
\newblock Reinforcement learning with unsupervised auxiliary tasks.
\newblock \emph{ICLR}, 2017{\natexlab{b}}.

\bibitem[Kempka et~al.(2016)Kempka, Wydmuch, Runc, Toczek, and
  Ja{\'s}kowski]{kempka2016vizdoom}
Kempka, Micha{\l}, Wydmuch, Marek, Runc, Grzegorz, Toczek, Jakub, and
  Ja{\'s}kowski, Wojciech.
\newblock Vizdoom: A doom-based ai research platform for visual reinforcement
  learning.
\newblock In \emph{Computational Intelligence and Games (CIG), 2016 IEEE
  Conference on}, pp.\  1--8. IEEE, 2016.

\bibitem[Li \& Yuan(2017)Li and Yuan]{conv1}
Li, Yuanzhi and Yuan, Yang.
\newblock Convergence analysis of two-layer neural networks with relu
  activation.
\newblock In \emph{Advances in Neural Information Processing Systems}, pp.\
  597--607, 2017.

\bibitem[Mirowski et~al.(2017)Mirowski, Pascanu, Viola, Soyer, Ballard, Banino,
  Denil, Goroshin, Sifre, Kavukcuoglu, et~al.]{mirowski2016learning}
Mirowski, Piotr, Pascanu, Razvan, Viola, Fabio, Soyer, Hubert, Ballard,
  Andrew~J, Banino, Andrea, Denil, Misha, Goroshin, Ross, Sifre, Laurent,
  Kavukcuoglu, Koray, et~al.
\newblock Learning to navigate in complex environments.
\newblock \emph{ICLR}, 2017.

\bibitem[Mnih et~al.(2016)Mnih, Badia, Mirza, Graves, Lillicrap, Harley,
  Silver, and Kavukcuoglu]{mnih2016asynchronous}
Mnih, Volodymyr, Badia, Adria~Puigdomenech, Mirza, Mehdi, Graves, Alex,
  Lillicrap, Timothy, Harley, Tim, Silver, David, and Kavukcuoglu, Koray.
\newblock Asynchronous methods for deep reinforcement learning.
\newblock In \emph{International Conference on Machine Learning}, pp.\
  1928--1937, 2016.

\bibitem[Parisotto et~al.(2016)Parisotto, Ba, and Salakhutdinov]{ParisottoBS15}
Parisotto, Emilio, Ba, Lei~Jimmy, and Salakhutdinov, Ruslan.
\newblock Actor-mimic: Deep multitask and transfer reinforcement learning.
\newblock \emph{ICLR}, 2016.

\bibitem[Ross et~al.(2011)Ross, Gordon, and Bagnell]{ross2011reduction}
Ross, St{\'e}phane, Gordon, Geoffrey, and Bagnell, Drew.
\newblock A reduction of imitation learning and structured prediction to
  no-regret online learning.
\newblock In \emph{Proceedings of the fourteenth international conference on
  artificial intelligence and statistics}, pp.\  627--635, 2011.

\bibitem[Rusu et~al.(2016)Rusu, Colmenarejo, Gulcehre, Desjardins, Kirkpatrick,
  Pascanu, Mnih, Kavukcuoglu, and Hadsell]{rusu-distillation-2015}
Rusu, Andrei~A, Colmenarejo, Sergio~Gomez, Gulcehre, Caglar, Desjardins,
  Guillaume, Kirkpatrick, James, Pascanu, Razvan, Mnih, Volodymyr, Kavukcuoglu,
  Koray, and Hadsell, Raia.
\newblock Policy distillation.
\newblock 2016.

\bibitem[Simonyan \& Zisserman(2014)Simonyan and Zisserman]{Simonyan14c}
Simonyan, K. and Zisserman, A.
\newblock Very deep convolutional networks for large-scale image recognition.
\newblock \emph{CoRR}, abs/1409.1556, 2014.

\bibitem[Sutskever \& Zaremba(2014)Sutskever and Zaremba]{Sutskever14}
Sutskever, Ilya and Zaremba, Wojciech.
\newblock Learning to execute.
\newblock \emph{CoRR}, 2014.

\bibitem[Teh et~al.(2017)Teh, Bapst, Czarnecki, Quan, Kirkpatrick, Hadsell,
  Heess, and Pascanu]{distral}
Teh, Yee, Bapst, Victor, Czarnecki, Wojciech~M., Quan, John, Kirkpatrick,
  James, Hadsell, Raia, Heess, Nicolas, and Pascanu, Razvan.
\newblock Distral: Robust multitask reinforcement learning.
\newblock In \emph{NIPS}. 2017.

\bibitem[Wei et~al.(2016)Wei, Wang, Rui, and Chen]{pmlr-v48-wei16}
Wei, Tao, Wang, Changhu, Rui, Yong, and Chen, Chang~Wen.
\newblock Network morphism.
\newblock In \emph{Proceedings of The 33rd International Conference on Machine
  Learning}, pp.\  564--572, 2016.

\bibitem[Zhang et~al.(2017)Zhang, Xiang, Hospedales, and Lu]{zhang2017deep}
Zhang, Ying, Xiang, Tao, Hospedales, Timothy~M, and Lu, Huchuan.
\newblock Deep mutual learning.
\newblock \emph{arXiv preprint arXiv:1706.00384}, 2017.

\end{thebibliography}
\bibliographystyle{icml2018}

\end{document}